\documentclass[jair,twoside,11pt,theapa]{article}
\usepackage{jair, theapa, rawfonts}

\usepackage[margin=1.2in]{geometry}

\usepackage{tablefootnote}

\usepackage{amsmath}
\usepackage{enumerate}

\usepackage{graphicx}
\usepackage{subcaption}
\usepackage[export]{adjustbox}
\usepackage{wrapfig}

\usepackage{booktabs}
\usepackage{array}
\usepackage[table]{xcolor}
\usepackage{multirow}

\usepackage{threeparttable}

\newcommand{\wrt}{{\it w.r.t. }}   
\newcommand{\eg}{\emph{e.g.}, }       
\newcommand{\ie}{\emph{i.e.}, }      
\newcommand\etc{\emph{etc.}}

\ShortHeadings{On the Behavior of Convolutional Nets for Feature Extraction}
{Garcia-Gasulla, Par\'{e}s, \& Vilalta}
\firstpageno{1}

\begin{document}

\title{On the Behavior of Convolutional Nets for Feature Extraction}

\author{\name Dario Garcia-Gasulla \email dario.garcia@bsc.es \\
       \name Ferran Par\'{e}s \email ferran.pares@bsc.es \\
       \name Armand Vilalta \email armand.vilalta@bsc.es \\
       \name Jonatan Moreno  \\
       \addr Barcelona Supercomputing Center (BSC)\\
       Omega Building, C. Jordi Girona, 1-3 08034 Barcelona, Catalonia
       \AND
       \name Eduard Ayguad\'{e} \\
       \name Jes\'{u}s Labarta \\
       \name Ulises Cort\'{e}s \\
       \addr Barcelona Supercomputing Center (BSC)\\
       \addr Universitat Polit\`{e}cnica de Catalunya - BarcelonaTech (UPC)
       \AND
       \name Toyotaro Suzumura  \\
       \addr IBM T.J. Watson Research Center, New York, USA\\
       \addr Barcelona Supercomputing Center (BSC)\\
       }


\maketitle

\begin{abstract}
Deep neural networks are representation learning techniques. During training, a deep net is capable of generating a descriptive language of unprecedented size and detail in machine learning. Extracting the descriptive language coded within a trained CNN model (in the case of image data), and reusing it for other purposes is a field of interest, as it provides access to the visual descriptors previously learnt by the CNN after processing millions of images, without requiring an expensive training phase. Contributions to this field (commonly known as feature representation transfer or transfer learning) have been purely empirical so far, extracting all CNN features from a single layer close to the output and testing their performance by feeding them to a classifier. This approach has provided consistent results, although its relevance is limited to classification tasks. In a completely different approach, in this paper we statistically measure the discriminative power of every single feature found within a deep CNN, when used for characterizing every class of 11 datasets. We seek to provide new insights into the behavior of CNN features, particularly the ones from convolutional layers, as this can be relevant for their application to knowledge representation and reasoning. Our results confirm that low and middle level features may behave differently to high level features, but only under certain conditions. We find that all CNN features can be used for knowledge representation purposes both by their presence or by their absence, doubling the information a single CNN feature may provide. We also study how much noise these features may include, and propose a thresholding approach to discard most of it. All these insights have a direct application to the generation of CNN embedding spaces.

\end{abstract}

\section{Introduction} \label{sec:i}

Image classification is among the most successful applications of deep learning. The performance of deep learning networks on challenges like the ImageNet Large Scale Visual Recognition Competition is based on the capabilities of these models at building an exceptionally rich representation language for a given dataset. A language that can be used for solving problems like classification or detection, achieving state-of-the-art performance \shortcite{he2016deep}. Coherently, deep learning models are frequently defined as representation learning techniques \shortcite{lecun2015deep}. Unfortunately, to build this representation language, deep networks require a lot of data and a lot of computational effort, which reduces the number of problems to which these models can be directly applied to. 

Within deep learning, the field of research commonly known as \textit{transfer learning} tries to reuse the representation language learnt for one problem to solve another. To formalize this we use the notation introduced by \shortciteA{pan2010survey}. This notation has two main components which can be summarized as follows; a domain $\mathcal{D}$ which is defined by a set of data instances with a given probability distribution (\eg images of a certain resolution with a certain distribution of pixel values), and a task $\mathcal{T}$ which is defined by a set of labels and a target function (\eg labels assigned to images, and a function classifying those images accordingly).


Transfer learning in deep learning is most frequently used to initialize a deep network using the features learnt for a source problem $(\mathcal{T_S},\mathcal{D_S})$, so that the same network can be optimized later through a fine-tuning process for a target problem $(\mathcal{T_T},\mathcal{D_T})$. Significantly, this approach has been shown to produce better results than training a network for $(\mathcal{T_T},\mathcal{D_T})$ from scratch (\ie using random initialization) \shortcite{yosinski2014transferable}. According to \shortciteA{pan2010survey}, this approach would be an example of \textit{inductive transfer learning}, since labeled data is available for both $\mathcal{T_T} $ and $ \mathcal{T_S}$. An alternative use of transfer learning is to use a neural network trained for $\mathcal{T_S}$ as a feature extractor for $\mathcal{T_T}$, in order to use other machine learning methods on top of the resulting representations. By doing so, one is representing the $\mathcal{D_T}$ data in a language learnt for the $\mathcal{T_S}$ task, enabling the use of pre-trained deep network representations (not deep network models by themselves) on datasets which lack the size required to train these methods \shortcite{azizpour2016factors,sharif2014cnn}. According to \shortciteA{pan2010survey} this case would be an example of \textit{feature representation transfer} (the analogous term \textit{transfer learning for feature extraction} is also widely used). Notice that this approach to transfer learning can be used to tackle unsupervised learning problems such as clustering \shortcite{gui2015learning}, and it would also enable the use of features obtained from an unsupervised learning task (\eg Autoencoders).



In the context of convolutional neural networks (CNNs), most attempts at feature representation transfer have focused on reusing the activations obtained from layers close to the output of the CNN (typically a fully-connected layer). When compared to lower-level layers, these high-level layers provide better results when used out-of-the-box to feed classifiers or clustering algorithms \shortcite{azizpour2016factors,sharif2014cnn,gui2015learning,donahue2014decaf}. Regardless of these results, all convolutional layers within a CNN encode a large amount of visual knowledge of varying complexity \shortcite{yosinski2015understanding,garcia2017visual}, knowledge which has not been successfully exploited so far. Since the general purpose of feature representation transfer is to maximize representativeness, we may hypothesize that the optimal representation will include, to some degree, information from a larger variety of layers (\ie beyond the last ones). It is therefore relevant, particularly for the knowledge representation and reasoning fields, to understand what differences are there between convolutional layers and fully connected layers, so that all that is learnt by a CNN can be properly exploited.

In this context, this paper analyzes the behaviour of all features within a deep CNN for the purpose of feature extraction. We use a very deep CNN (\texttt{VGG16}, \citeR{simonyan2014very}) pre-trained on a large dataset ($(\mathcal{T_S},\mathcal{D_S})$ = \textit{ImageNet 2012}) to build image representations for alternative datasets ($(\mathcal{T_T},\mathcal{D_T}) \in \{mit67, flowers102, cub200,...\}$), and study the behavior of individual features for the domain defined by each dataset. First, Section \ref{sec:rw} introduces previous contributions to the transfer learning field, focusing on feature extraction. Section \ref{sec:dmabe} introduces the datasets and CNN model used in our experiments, as well as the image embedding we build in our feature extraction process. The basis of our study are statistical distance methods, which we review in Section \ref{sec:sdm}. Section \ref{sec:sda}, introduces the behavior of statistical distances for the current problem, and the distributions these compose. These distributions are analyzed in Section \ref{sec:sdb}, while the impact of noise on the measures is discussed in Section \ref{sec:apdpc}. The consistency of our findings given a different source task is shown in Section \ref{sec:places}. Finally, the conclusions drawn from this study are summarized in Section \ref{sec:c}.

\section{Related Work} \label{sec:rw}

CNN models are defined by a large number of parameters, requiring lots of data instances (typically images) for their optimization. Until the release of large visual datasets, hand-made features \shortcite{perronnin2010improving} produced the best results for vision tasks. Nowadays, CNNs can be trained using datasets such as \textit{ImageNet} and \textit{VOC2012} \shortcite{pascalvoc2012}, learning powerful visual descriptors, which allows them to outperform previously competitive solutions such as Improved Fisher Vectors on many visual tasks \shortcite{chatfield2014return}.

\shortciteA{donahue2014decaf} presents one of the first studies on the behavior of convolutional filters in a feature extraction process. In that work, authors study a CNN (AlexNet architecture composed by 5 convolutional layers and 3 fully connected layers) trained using \textit{ImageNet 2012} as $(\mathcal{T_S},\mathcal{D_S})$ for transfer learning. Qualitatively, authors observe how features from the first fully connected layer outperform features from lower layers at the task of separating concepts according to the WordNet hierarchy. Authors evaluate features extracted from the last convolutional layer and the first two fully connected layers on various datasets. Their results indicate that by using features from the first fully connected layers to train a support vector machine (SVM) one can achieve state-of-the-art results on various related tasks.

The contribution of \shortciteA{sharif2014cnn} goes in a similar direction, using the OverFeat network architecture pre-trained using \textit{ImageNet 2012} as $(\mathcal{T_S},\mathcal{D_S})$. Authors focus mostly on the first fully connected layer, performing data augmentation to increase the quality of those features (cropping and rotating samples), and doing component-wise power transformation. After applying a l2-normalization to the resultant vectors, an SVM is trained and applied to a wide variety of tasks ($\mathcal{T_T} \in $\{image classification, fine grained recognition, attribute detection, visual instance retrieval,...\}) and domains ($\mathcal{D_T} \in$ \{\textit{VOC2007, flowers102, cub200,...}) achieving competitive results on all of them. For one of those tasks (image classification), features from various layers are evaluated separately using an SVM, with the first fully connected layer obtaining the best results.

A rather different approach is depicted by \shortciteA{yosinski2014transferable}, where the goal is to study the transferability of features for the purpose of fine tuning the deep neural network for the target task and dataset. In that regard, authors find that the distance between the source and target tasks is strongly related with the depth of the optimal layer to be used in the transfer learning process.


\shortciteA{azizpour2016factors} empirically evaluated several parameters that can affect the transfer learning process for feature extraction. Among the parameters they considered some are related with the architecture and training of the initial CNN (network depth and width, distribution of training data, optimization parameters), and some are related with the transfer learning process (fine-tuning, network layer to be extracted, spatial pooling and dimensionality reduction). All these parameters are evaluated on 17 visual recognition tasks, identifying a set of good parameters depending on the distance between the source task and the target task. Regarding the representation layer (which layer is used to build the embedding) authors find that the first or second fully connected layer produces the best results on most cases, when feeding an SVM for classification.

Deep residual networks (ResNets) are an evolution of traditional CNN which include branching of paths. Unlike CNNs, which stack layers sequentially, ResNets implement shortcut connections which eases the convergence during the training process, allowing the training of networks with more layers (up to thousands). \shortciteA{mahmood2016resfeats} explore the use of ResNets for feature extraction, particularly to solve three image classification problems. Results indicate that ResNets are a competitive alternative to classic CNN architectures, also in the context of feature extraction.


\shortciteA{long15} proposed a new Deep Adaptation Network (DANs) architecture to solve the problem of domain adaptation for convolutional neural networks. In this architecture the first convolutional layers parameters are reused without modification, while the weights of the last convolutional layers are fine-tuned for the new task. Fully connected weights are tailored to fit specific tasks via Multiple Kernel Maximum Mean Discrepancies (MK-MMD). However, this approach does not consider the problem of solving a different target task and its impact in the reusability of the pre-trained features.

\section{Source and Target Problems}  \label{sec:dmabe}


Feature representation transfer in the context of deep learning is particularly useful when the target problem $(\mathcal{T_T},\mathcal{D_T})$ does not include enough labeled data to train its own CNN representation language. In this context, the $(\mathcal{T_T},\mathcal{D_T})$ problem can be solved by using features crafted for a different $(\mathcal{T_S},\mathcal{D_S})$ problem, although the quality of the resultant embedding representation will strongly depend on how similar ($\mathcal{D_S}$,$\mathcal{T_S}$) and ($\mathcal{D_T}$,$\mathcal{T_T}$) are. If the language learnt for $(\mathcal{T_S},\mathcal{D_S})$ lacks the vocabulary to properly characterize the particularities of $(\mathcal{T_T},\mathcal{D_T})$ (\eg because $\mathcal{D_S}$ is defined by black and white images, and $\mathcal{D_T}$ includes colorful patterns), the resultant embedding will be of poor quality and any learning applied to it will be deficient. With that in mind, $(\mathcal{T_S},\mathcal{D_S})$ is typically chosen to capture a range of visual patterns as broad as possible, so that its language will be likely to include features relevant for many different target tasks, and capable of characterizing a wide variety of image domains. In this regard, a CNN trained for the \textit{ImageNet 2012} dataset \shortcite{ILSVRC15} is a good candidate for a source problem, since the 1,000 categories composing its $\mathcal{T_S}$ requires a huge variety of visual patterns, while the large number of images available in $\mathcal{D_S}$ guarantees that the learnt model will generalize to different domains.

\subsection{CNN Architecture}\label{sec:arch}

To completely specify $\mathcal{T_S}$ one needs the label space $\mathcal{Y}$ but also an objective predictive function $f(\cdot)$. In our case the function $f(\cdot)$ is defined by a trained CNN (\ie its architecture and parameters). There are many popular CNN architectures, and various have been used for feature extraction (see Section\ref{sec:rw}). Since our goal is to explore the behavior of convolutional layers in the feature extraction process, we will use an architecture which follows the most canonical scheme of layers (\ie \texttt{conv/pool/conv/pool/.../fc}). At the same time, we wish to use a model capable of learning a rich representation language at various levels (\ie a very deep network). This combination of requirements leads us to use the \texttt{VGG16} architecture as source of features \shortcite{simonyan2014very}. \texttt{VGG16} is composed by 13 convolutional layers (with 5 pooling layers) and 3 fully-connected layers (see Table \ref{tab:vgg16} for details on the architecture). The only exception are Figures \ref{fig:topKSimg_mit67}, \ref{fig:botKSimgcub200}, \ref{fig:botKSimgflowers102} and \ref{fig:topKconv1flowers102} which are obtained using the \texttt{VGG19} architecture, and used here only for illustrative purposes. This architecture is from the same authors, and detailed on the same paper. It only differs from \texttt{VGG16} by having 3 extra convolutional layers \texttt{conv3\_4, conv4\_4 and conv5\_4}. Results obtained with the \texttt{VGG19} architecture were consistent with the ones obtained with \texttt{VGG16} for all experiments. Both models are publicly available at the authors web page\footnote{http://www.robots.ox.ac.uk/$\sim$vgg/research/very\_deep/}.

\begin{table}[b!]
\centering
\caption{Details of the \texttt{VGG16} architecture \shortcite{simonyan2014very}. For each layer: number of filters (\ie neurons with unique set of parameters), learnable parameters (\ie weights and biases), neurons and features in the representation transfer process.}
\label{tab:vgg16}
\begin{tabular}{lrrrr}
\toprule
Layer name     & \#Filters   & \#Parameters & \#Neurons & \# Embedding Features   \\ 
\midrule
\texttt{input}          &             & 150K         &      & -             \\
\midrule
\texttt{conv1\_1}       & 64          & 1.7K         & 3.2M & 64    \\ 
\texttt{conv1\_2}       & 64          & 36K          & 3.2M & 64    \\ 
\texttt{pool1}      &\multicolumn{2}{c}{max pooling} & 802K & -     \\
\midrule
\texttt{conv2\_1}       & 128         & 73K          & 1.6M & 128   \\ 
\texttt{conv2\_2}       & 128         & 147K         & 1.6M & 128   \\ 
\texttt{pool2}      &\multicolumn{2}{c}{max pooling} & 401K & -     \\ 
\midrule
\texttt{conv3\_1}       & 256         & 300K         & 802K & 256   \\ 
\texttt{conv3\_2}       & 256         & 600K         & 802K & 256   \\ 
\texttt{conv3\_3}       & 256         & 600K         & 802K & 256   \\
\texttt{pool3}      &\multicolumn{2}{c}{max pooling} & 200K & -     \\ 
\midrule
\texttt{conv4\_1}       & 512         & 1.1M         & 401K & 512   \\ 
\texttt{conv4\_2}       & 512         & 2.3M         & 401K & 512   \\ 
\texttt{conv4\_3}       & 512         & 2.3M         & 401K & 512   \\  
\texttt{pool4}      &\multicolumn{2}{c}{max pooling} & 100K & -     \\  
\midrule
\texttt{conv5\_1}       & 512         & 2.3M         & 100K & 512   \\  
\texttt{conv5\_2}       & 512         & 2.3M         & 100K & 512   \\  
\texttt{conv5\_3}       & 512         & 2.3M         & 100K & 512   \\ 
\texttt{pool5}      &\multicolumn{2}{c}{max pooling} & 25K  & -     \\  
\midrule
\texttt{fc6}            & 4,096       & 103M         & 4K   & 4,096 \\  
\texttt{fc7}            & 4,096       & 17M          & 4K   & 4,096 \\   
\midrule
\texttt{output}         & 1,000       & 4M           & 1K   & -     \\
\midrule
\textbf{Total}          & \textbf{13,416}      & \textbf{138M}         & \textbf{15M}  & \textbf{12,416} \\
\bottomrule
\end{tabular}
\end{table}

\subsection{Target Datasets}\label{sec:dts}

Once we have defined the source task and domain $(\mathcal{T_S},\mathcal{D_S})$, let us introduce the publicly available datasets we will consider as target $(\mathcal{T_T},\mathcal{D_T})$ in our study on transfer learning:
\begin{enumerate}
    \item The MIT Indoor Scene Recognition dataset \shortcite{quattoni2009recognizing} (\textit{mit67}) consists of different indoor scenes of 67 categories. Its main challenge resides in the class dependence on global spatial properties and on the relative presence of objects.
    \item The Caltech-UCSD Birds-200-2011 dataset \shortcite{WahCUB_200_2011} (\textit{cub200}) is a fine-grained dataset containing images of 200 different species of birds.
    \item The Oxford Flower dataset \shortcite{nilsback2008automated} (\textit{flowers102}) is a fine-grained dataset consisting of 102 flower categories.
    \item The Oxford-IIIT-Pet dataset \shortcite{parkhi2012cats} (\textit{catsdogs}) is a fine-grained dataset covering 37 different breeds of cats and dogs.
    \item The Stanford Dogs dataset \shortcite{khosla2011novel} (\textit{stanforddogs}) another fine-grained dataset containing images from the 120 breeds of dogs found in \textit{ImageNet 2012}.
    \item The Caltech-101 dataset \shortcite{fei2007learning} (\textit{caltech101}) is a classical dataset of 101 object categories containing clean images with low level of occlusion.
    \item The Caltech-256 dataset \shortcite{griffin2007caltech} (\textit{caltech256}) contains 256 object categories with a larger minimum number of images per class than \textit{caltech101}.
    \item The Food-101 dataset \shortcite{bossard2014food} (\textit{food101}) is a large dataset of 101 popular food categories. In this particular case we only use the test split of the data to reduce its size and avoid the label-noise present in the train split.
    \item The Describable Textures Dataset \shortcite{cimpoi2014describing} (\textit{textures}) is a database of textures categorized according to a list of 47 terms inspired from human perception.
    \item The Oulu Knots dataset \shortcite{silven2003wood} (\textit{wood}) contains knot images from spruce wood, classified according to Nordic Standards. This old dataset of industrial application is considered to be challenging even for human experts.
    \item We also use the validation split of \textit{ImageNet 2012} \shortcite{ILSVRC15} (\textit{imagenet}) as a target problem for comparison purposes. Notice that the images composing this dataset are different from the training set of \textit{ImageNet 2012}, so they can have a different distribution which implies that domains can be different, but similar ($D_S \simeq D_T$). For disambiguation we will refer to \textit{ImageNet 2012} when talking about the whole dataset and to \textit{imagenet} when talking about this target task $\mathcal{T_T}$.
\end{enumerate}
          
Dataset sizes, number of classes and number of images per class are specified in Table \ref{tab:datasets}. In our experiments we do not train models using these datasets, which means we do not require the provided train and test splits. Instead, we will merge both splits to make use of all the data available.

\begin{table}[t]
    \caption{Properties of datasets used in our experiments}
    \label{tab:datasets}
    \centering
    \begin{tabular}{lrrrr}
        \toprule
                    &           &           &       & {\centering Average} \\
        Dataset     & \#Images  & \#Classes &  \#Images per class    & \#Images per class     \\
                    &  $|I|$    & $|C|$     &  $|I_c|$      & $\overline{|I_c|}$\\
        \midrule
        \textit{imagenet}    &   50,000  &   1000    &   50          & 50            \\
        \textit{cub200}      &   11,788  &   200     &   41 - 60     & 59            \\
        \textit{wood}        &   438     &   7       &   14 - 179    & 63            \\
        \textit{flowers102}  &   8,189   &   102     &   40 - 258    & 80            \\
        \textit{caltech101}  &   9,146   &   101     &   31 - 800    & 91            \\
        \textit{mit67}       &   6,700   &   67      &   100         & 100           \\ 
        \textit{textures}    &   5,640   &   47      &   120         & 120           \\
        \textit{caltech256}  &   30,607  &   256     &   80 - 800    & 120           \\
        \textit{stanforddogs}&   20,580  &   120     &   150 - 200   & 172           \\
        \textit{catsdogs}    &   7,349   &   37      &   184 - 200   & 199           \\
        \textit{food101}     &   25,250  &   101     &   250         & 250           \\
        \bottomrule
    \end{tabular}
\end{table}

\subsection{Embedding}\label{sec:emb}

Given the source and target problems $(\mathcal{T_S},\mathcal{D_S}),(\mathcal{T_T},\mathcal{D_T})$, there are still several parameters than can modify the construction of the embedding space. Most of those parameters were explored by \shortciteA{azizpour2016factors}. In our case, we will use two main parameters which we consider to be coherent with our study. First, each image representation will be built as a result of processing 10 crops of the image (4 corners and middle crop, mirrored) through the CNN and averaging the resulting activations. This is a frequently used methodology for feature extraction \shortcite{sharif2014cnn,azizpour2016factors}, which provides robustness to the resultant representations. Second, we perform a spatial average pooling of each convolutional layer to obtain a single value per filter. This transformation reduces the number of features in the embedding, as well as the relative spatial information (\ie each resulting feature will determine if a visual pattern is found or not in the image on average, regardless of its exact location), while maintaining most of its descriptive power (\ie each feature is still separately accounted for in the embedding). This spatial pooling methodology is also a recurrent solution in the field \shortcite{sharif2014cnn,azizpour2016factors}.

Since we wish to explore the behavior of convolutional layers, our embedding will contain all the 16 convolutional layers available in \texttt{VGG16} (from \texttt{conv1\_1} to \texttt{conv5\_3}). For comparison purposes we will also extract the fully connected layers (\texttt{fc6,fc7}), so that we can contrast the behavior of the convolutional and fully connected features. Notice the spatial pooling performed on the convolutional layers cannot be applied to the \texttt{fc} layers. The components of the resultant embedding, composed by 12,416 values, is shown in Table \ref{tab:vgg16}. For the remaining of the document, all mentions to \textit{the embedding} will refer to this representation.



\section{Statistic Distance Methods} \label{sec:sdm}

Previous studies on the usefulness of convolutional layers for feature extraction transfer learning have been purely empirical, based on the performance of specific classifiers (most frequently, an SVM) using the features extracted from a single layer of a CNN (see Section \ref{sec:rw}). This approach has been shown to provide consistent results, but it is limited to classification, and strongly influenced by the choice of classifier (\eg some classifiers may perform better with a certain number of variables, or may be affected differently by noise). 

In this paper we propose a different approach to evaluate the behavior of CNN features. Instead of evaluating the performance of a specific machine learning algorithm on the embedding, we measure the descriptive power of CNN features statistically, studying their behavior for the different classes composing each dataset. The goal is to learn about the descriptive nature of CNN features, so that other knowledge representation and reasoning methodologies can be adapted accordingly.

In detail, our approach consists on evaluating how characteristic each feature in the embedding is, for each of the target classes of the considered datasets. In other words, we do not want to evaluate the descriptive power of a group of features (which would be a feature selection problem) but to analyze the discriminative power of each single feature. CNN neurons do not have a crisp behavior \wrt classes (\ie neurons do not activate binarily depending on the class), not even for the original training task. Instead, each CNN neuron provides a fuzzy piece of information for each class. To contextualize the information provided by individual features, we consider their activations on a given class of the target task $\mathcal{T_T}$ (\textit{inner-class behaviour}), and compare it with the activations happening for the rest of the classes of the same dataset $\mathcal{T_T}$ (\textit{outer-class behaviour}). This will also give us an insight on how these features would perform on their own for representing each single class within a dataset.

\begin{figure}[t!]
    \centering
    \begin{tabular}{cc}
        \includegraphics[width=0.47\linewidth]{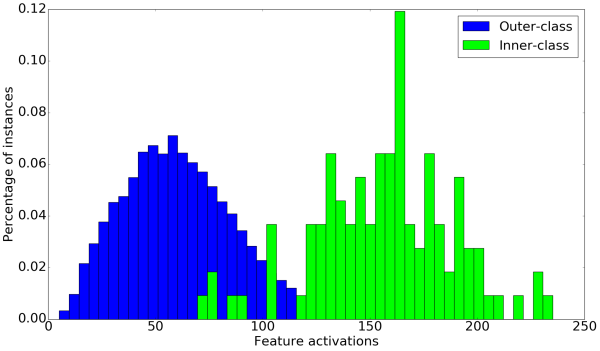} &
        \includegraphics[width=0.47\linewidth]{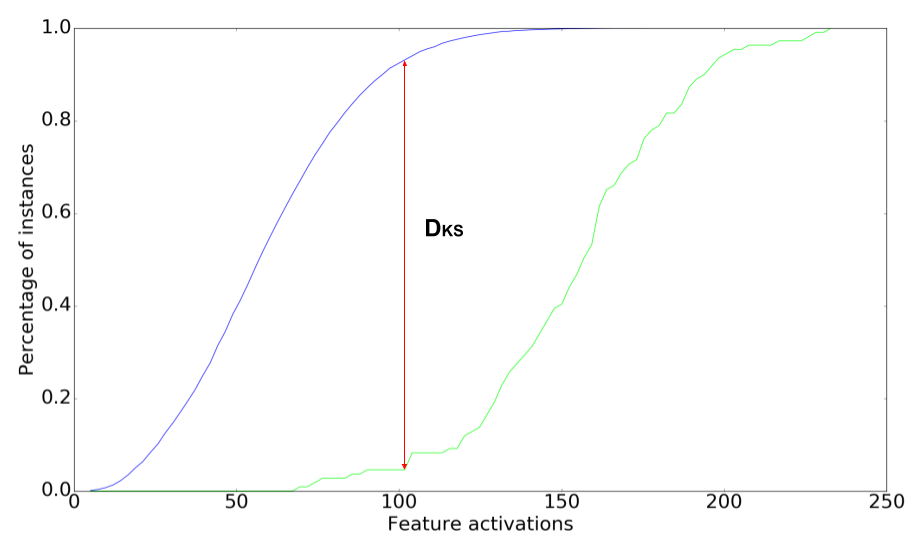}\\       
    \end{tabular}
    \vspace{-10pt}
    \caption{Left: Inner/outer class density estimations for feature \texttt{n458} from layer \texttt{conv4\_3}, for class \textit{pink-yellow dahlia} (60) of the \textit{flowers102} dataset. Right: Accumulated distributions and Kolmogorov-Smirnov statistic ($D_{KS}$) for the same data. Right line indicates maximum distance, corresponding to the Kolmogorov-Smirnov statistic.}
    \label{fig:inner_inter_hist}
\end{figure}

The inner/outer class behaviour can be visualized through two histograms of feature activations (see left plot of Figure \ref{fig:inner_inter_hist}). Statistically speaking, rescaled histograms are density estimations approximating a true probability density function (PDF). Although more sophisticated methods are available \shortcite{scott2015multivariate} we use the histogram for the sake of computational simplicity. To study the behavior of a given CNN feature for a given class (a feature class pair), we compare the corresponding inner/outer density estimations. The first statistical distance we consider using for that purpose is the well-known Kullback-Leibler (KL) divergence.

The Kullback-Leibler divergence measures how much two PDF, $P$ and $Q$, differ following Equation \ref{eq:kl}, where $i$ are the points in the domain.
\begin{equation} \label{eq:kl}
    D_{KL}(P,Q) = \sum_{i}P(i)ln\frac{P(i)}{Q(i)}
\end{equation}
Although histograms are only approximations of PDFs, it is possible to fit a PDF (\eg normal distribution, uniform distribution, etc) to a histogram. This is, however, inconvenient since the histogram of different features may be fit by different PDFs, and there may be some features which are not properly fitted by a PDF.

Mutual information measures the information that two random variables, $X$ and $Y$, share. It can be understood as the expectation of the Kullback–Leibler divergence of the univariate distribution $p(x)$ from the conditional distribution $p(x|y)$. The information gain is greater as the difference between the distributions $p(x|y)$ and $p(x)$ grows. In our experiments, $Y$ is analogous to \textit{belonging to class $c$}. Thus, $p(x|y)$ represents the inner-class distribution, and $p(x|\neg y)$ represents the outer-class distribution. This $p(x|\neg y)$ can be an approximation of $p(x)$ if the number of samples of other classes is much bigger than for class $c$, formally $|I_{\neg c}| \gg |I_c| \Rightarrow p(x|\neg y) \simeq p(x)$. In this case, which is the usual for tasks with high number of classes and evenly distributed samples, the mutual information can be approximated by the Kullback-Leibler divergence.

The Bhattacharyya distance is an alternative to KL which can measure the distance between two discrete probability distributions. Analogously, it can be measured from two density estimations $P$ and $Q$ following Equation \ref{eq:b}, where $i$ are the discrete points of the domain $X$. 
\begin{equation} \label{eq:b}
    D_{B}(P,Q) = -ln\left ( \sum_{i \in X}\sqrt{P(i)Q(i)} \right )
\end{equation}
By comparing two density estimations, the Bhattacharyya distance can be used directly on the data, without having to choose a fitting PDF. However, its mathematical range is only positive ($[0,\infty)$), making the Bhattacharyya distance unable to identify which density estimation is \textit{above} and which is \textit{below}. In our analysis it will be of interest to know if an inner-class behaviour is \textit{higher} than the outer-class behavior or vice versa, since both situations may provide different insights.

The Kolmogorov-Smirnov statistic ($D_{KS}$) measures the distance between two empirical distribution functions (EDF) $P$ and $Q$. For each point $i$ in the domain $X$, $D_{KS}$ evaluates the distance between $P(i)$ and $Q(i)$, and obtains the maximum. It is formally defined in Equation \ref{eq:ks} and graphically displayed in the right plot of Figure \ref{fig:inner_inter_hist}. To reduce the computational cost of evaluating $D_{KS}$, we discretize the domain of each EDF into 100 bins, thus decreasing the domain resolution by $1\%$. Notice that, since EDF is a cumulative distribution, $D_{KS}$ is directly computed on a set of values (one pair of inner/outer class behaviours).

\begin{equation} \label{eq:ks}
    D_{KS}(P,Q) = \sup_{i\in X} | P(i) - Q(i) |
\end{equation}

In contrast with Bhattacharyya distance, $D_{KS}$'s mathematical range is $[0,1]$. We use a signed variant of $D_{KS}$ where the sign indicates which EDF is above and which is below at the point where $P$ and $Q$ differ most. This variant extends the range of $D_{KS}$ to $[-1,1]$ and allows us to differentiate when inner class behavior is above outer class behavior ($D_{KS} > 0$) and vice versa ($D_{KS} < 0$). $D_{KS}=0$ means that both distributions are identical, while $D_{KS}=1$ and $D_{KS}=-1$ means that both distributions do not intersect at any point. The Kolmogorov-Smirnov statistic does not require the fitting of a PDF (unlike Kullback-Leibler divergence) which is desirable. For these reasons, in all our following experiments we will use the signed version of Kolmogorov-Smirnov statistic ($D_{KS}$).


\section{Statistic Distance Behaviors} \label{sec:sda}

Our statistic distance analysis is based on the inner/outer class $D_{KS}$. In this section we introduce the behavior of the $D_{KS}$ values, and the distribution of these values layer-wise when computing a given datasets. The following section \ref{sec:dist_dist} contains a detailed study of the distributions from various perspectives.

Simply put, a distance $D_{KS}(f,c) \simeq 0$ means that the distribution of activations of feature $f$ for all the images belonging to class $c$ (\ie $I_c$) is almost identical to the distribution of values of feature $f$ for all the images that do not belong to that class (\ie $I_{\neg c}$). If $D_{KS}(f,c) > 0$ then feature values for images $I_c$ tend to be higher than for the rest of images $I_{\neg c}$, which implies that the visual elements represented by feature $f$ are more commonly found in $I_c$ images than in $I_{\neg c}$ images. Similarly, if $D_{KS}(f,c) < 0$, feature values are in general lower for $I_c$ than for $I_{\neg c}$, which implies that elements represented by feature $f$ are rare within $I_c$ images when compared to the rest of the dataset.
\begin{figure}[t!]
    \renewcommand{\arraystretch}{0}
    \centering
    \begin{tabular}{c *{4}{m{0.2\textwidth}}}
        &
        \begin{center}\texttt{conv3\_4 n202}\end{center}&
        \begin{center}\texttt{conv4\_2 n89}\end{center}&
        \begin{center}\texttt{conv5\_4 n277}\end{center}&
        \begin{center}\texttt{fc7 n1426}\end{center}\\
        
        \vspace{-10pt}\parbox[c]{2mm}{\rotatebox[origin=c]{90}{\phantom{some }f visualization}}&
        \vspace{-10pt}\begin{center}\includegraphics[width=1\linewidth]{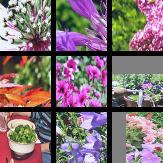}\end{center}&
        \vspace{-10pt}\begin{center}\includegraphics[width=1\linewidth]{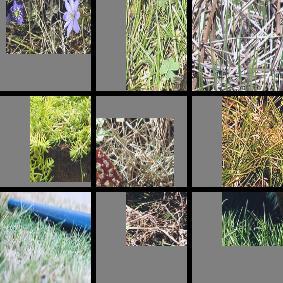}\end{center}&
        \vspace{-10pt}\begin{center}\includegraphics[width=1\linewidth]{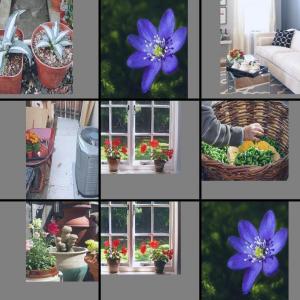}\end{center}&
        \vspace{-10pt}\begin{center}\includegraphics[width=1\linewidth]{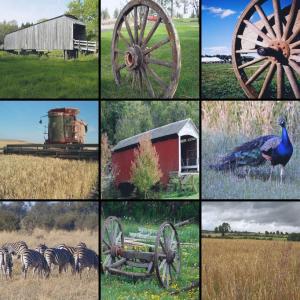}\end{center}\\
        
        \vspace{-10pt}
        class &  
        \vspace{-0pt}\begin{center}Greenhouse\end{center}&
        \vspace{-0pt}\begin{center}Greenhouse\end{center}&
        \vspace{-0pt}\begin{center}Florist\end{center}&
        \vspace{-0pt}\begin{center}Greenhouse\end{center}\\

        $D^+_{KS}$&
        \vspace{-0pt}\begin{center}$ 0.9144 $\end{center}&
        \vspace{-0pt}\begin{center}$ 0.9059 $\end{center}&
        \vspace{-0pt}\begin{center}$ 0.9287 $\end{center}&
        \vspace{-0pt}\begin{center}$ 0.9515 $\end{center}\\
        
        &
        \vspace{-0pt}\begin{center}\texttt{conv3\_3 n145}\end{center}&
        \vspace{-0pt}\begin{center}\texttt{conv4\_3 n293}\end{center}&
        \vspace{-0pt}\begin{center}\texttt{conv5\_4 n449}\end{center}&
        \vspace{-0pt}\begin{center}\texttt{fc7 n1779}\end{center}\\
        
        \vspace{-10pt}\parbox[c]{2mm}{\rotatebox[origin=c]{90}{\phantom{some }f visualization}}&
        \vspace{-10pt}\begin{center}\includegraphics[width=1\linewidth]{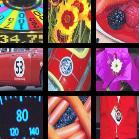}\end{center}&
        \vspace{-10pt}\begin{center}\includegraphics[width=1\linewidth]{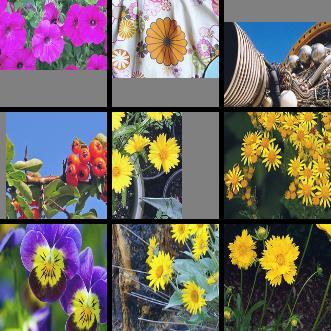}\end{center}&
        \vspace{-10pt}\begin{center}\includegraphics[width=1\linewidth]{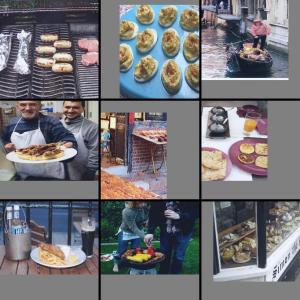}\end{center}&
        \vspace{-10pt}\begin{center}\includegraphics[width=1\linewidth]{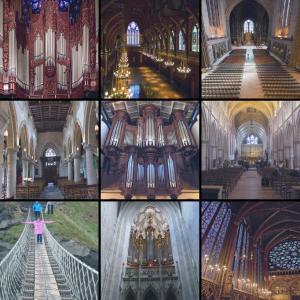}\end{center}\\  
        
        \vspace{-10pt}
        class&  
        \vspace{-0pt}\begin{center}Florist\end{center}&
        \vspace{-0pt}\begin{center}Florist\end{center}&
        \vspace{-0pt}\begin{center}Buffet\end{center}&
        \vspace{-0pt}\begin{center}Cloister\end{center}\\
        
        $D^+_{KS}$&
        \vspace{-0pt}\begin{center}$ 0.8641 $\end{center}&
        \vspace{-0pt}\begin{center}$ 0.8923 $\end{center}&
        \vspace{-0pt}\begin{center}$ 0.9259 $\end{center}&
        \vspace{-0pt}\begin{center}$ 0.9174 $\end{center}\\
        
    \end{tabular}
    \vspace{-0pt}
    \caption{8 embedding features with very high $D_{KS}$ (among top 10) for the \textit{mit67} dataset and \texttt{VGG19} architecture. The class producing that high $D_{KS}$ is shown below each feature. Each feature corresponds to a specific neuron (for fully connected layers) or filter (for convolutional layers) from the original CNN model. To illustrate captured visual patterns of each feature, we show 9 cropped images from \textit{ImageNet 2012} validation set producing the highest activation values for this neuron or filter. Images are cropped to match the neuron receptive field.}
    \label{fig:topKSimg_mit67}
\end{figure}

To illustrate this behavior we explore which are the features with the highest $D_{KS}$ values for the \textit{mit67} dataset (\ie closer to 1). Figure \ref{fig:topKSimg_mit67} shows some of the top $D_{KS}(f,c)$ values for different layers, indicating the class $c$ in which the large $D_{KS}$ value occurs. To show what a particular feature is encoding, we plot the 9 image crops from the \textit{ImageNet 2012} validation set (\ie \textit{imagenet}) producing the highest activation value for that feature. Images from this dataset will provide better feature characterizations, since CNN features were originally trained for its classes. In the example of Figure \ref{fig:topKSimg_mit67}, features having a high $D_{KS}$ value for the \textit{Greenhouse} class are either showing plants, grass or fields, regardless of the layer depth. The feature with a high $D_{KS}$ value for the \textit{Buffet} class identifies food on plates, while the feature with a high $D_{KS}$ value for the \textit{Cloister} identifies Gothic arches. In the case of feature \texttt{conv3\_3 n145}, which presents a high $D_{KS}$ value for the \textit{Florist} class, the crops producing high activations correspond to colorful patterns in contrast with its surroundings.

\begin{figure}[b!]
    \centering
    \begin{tabular}{c *{4}{m{0.18\textwidth}}}
        &
        \multicolumn{4}{c}{\texttt{fc7 n1946}}\\
        \vspace{-115pt}
        \parbox[c]{2mm}{\rotatebox[origin=c]{90}{\hphantom{some invisible text here, damn} Feature visualization}}&
        \multicolumn{4}{c}{\includegraphics[width=.3\linewidth]{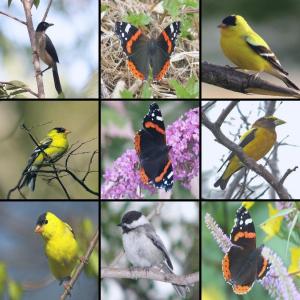}}\\
        \vspace{-15pt}
        \parbox[t]{2mm}{\rotatebox[origin=c]{90}{sample}}&
        \begin{center}\includegraphics[width=0.8\linewidth]{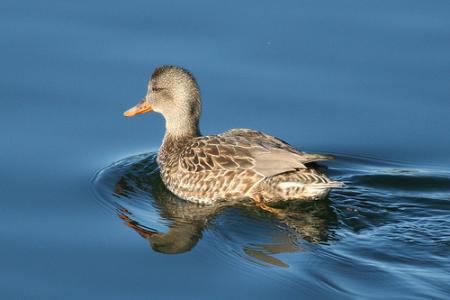}\end{center}&
        \begin{center}\includegraphics[width=0.8\linewidth]{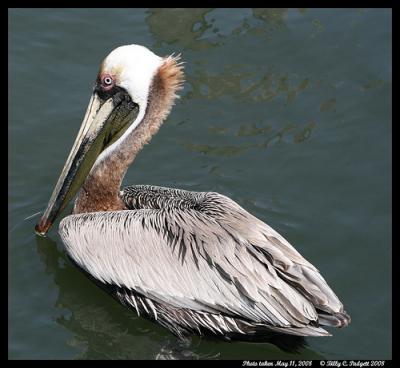}\end{center}&
        \begin{center}\includegraphics[width=0.8\linewidth]{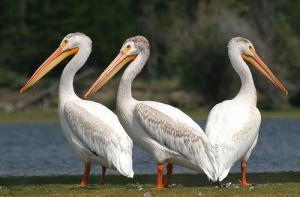}\end{center}&
        \begin{center}\includegraphics[width=0.8\linewidth]{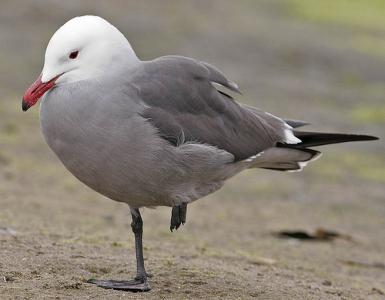}\end{center}\\ 
        \vspace{-15pt}
        class&           
        \begin{center}Gadwall\end{center}&           
        \begin{center}Brown Pelican\end{center}&
        \begin{center}White Pelican\end{center}&
        \begin{center}Heermann Gull\end{center}\\
        $D^-_{KS}$&
        \begin{center}$-0.8880$\end{center}&
        \begin{center}$-0.8736$\end{center}&
        \begin{center}$-0.8624$\end{center}&
        \begin{center}$-0.8553$\end{center}\\
        
    \end{tabular}
    \vspace{-10pt}
    \caption{Example of a filter with an extremely negative $D_{KS}$ value for four different classes of the \textit{cub200} dataset and \texttt{VGG19} architecture. First row illustrates the visual pattern being captured by the feature, using the same method as in Figure \ref{fig:topKSimg_mit67}. The second and third row contain sample images from the \textit{cub200} dataset for the four different classes, and the name of those classes.}
    \label{fig:botKSimgcub200}
\end{figure}

Analogous to the study of positive $D_{KS}$ values of Figure \ref{fig:topKSimg_mit67}, we consider the lowest $D_{KS}$ values (\ie closer to -1). Initially, one could expect that the features having the lowest $D_{KS}$ for a given class $c$ would be those identifying elements which never appear in the images of $c$. For example, a hypothetical class \textit{whale} could be expected to have a extremely negative $D_{KS}$ for a feature identifying a wheel. However, since the $D_{KS}$ values are computed in the context of a dataset (\ie it indicates inner/outer class disparity) such an assumption is incomplete. As a matter of fact, features having the lowest $D_{KS}$ for a given class $c$ are those identifying elements which appear in the images of $c$ very rarely \textit{when compared} with their frequency for the rest of images. For example, in a dataset composed only by the classes \textit{whale} and \textit{clownfish}, the features with the lowest $D_{KS}$ values for the class \textit{whale} would correspond to those having the highest $D_{KS}$ values for the class \textit{clownfish}, most likely features identifying orange related patterns. On the other hand, features identifying uncommon patterns on both classes (\ie a wheel) would have a $D_{KS}$ value close to zero for both classes, as its inner/outer class distributions would be very similar.

\begin{figure}[t!]
    \centering
    \begin{tabular}{c m{0.15\textwidth} m{0.15\textwidth} m{0.15\textwidth}| m{0.15\textwidth} m{0.15\textwidth}}
        &
        \multicolumn{3}{c|}{\texttt{fc7 n1449}}&
        \multicolumn{2}{c}{\texttt{fc7 n3529}} \vspace{-15pt}\\
        \vspace{-95pt}
        \parbox[t]{2mm}{\rotatebox[origin=c]{90}{\hphantom{some invisible text here}Feature visualization}}&
        \multicolumn{3}{c|}{\includegraphics[width=.20\linewidth]{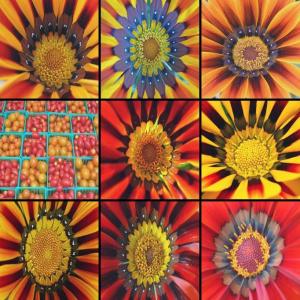}}&
        \multicolumn{2}{c}{\includegraphics[width=.20\linewidth]{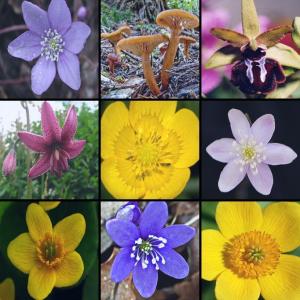}}\\
        \vspace{-15pt}
        \parbox[t]{2mm}{\rotatebox[origin=c]{90}{sample}}&
        \begin{center}\includegraphics[width=0.8\linewidth]{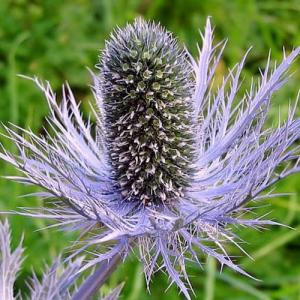}\end{center}&
        \begin{center}\includegraphics[width=0.8\linewidth]{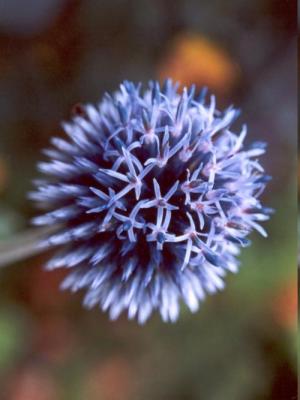}\end{center}&
        \begin{center}\includegraphics[width=0.8\linewidth]{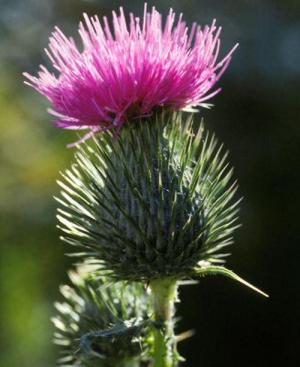}\end{center}&
        \begin{center}\includegraphics[width=0.8\linewidth]{flowers14}\end{center}&
        \begin{center}\includegraphics[width=0.8\linewidth]{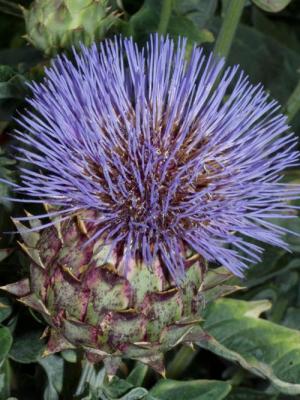}\end{center}\\ 
        
        \vspace{-15pt}
        class&           
        \begin{center}35 alpine sea holly\end{center}&           
        \begin{center}10 globe thistle\end{center}&
        \begin{center}14 spear thistle\end{center}&
        \begin{center}14 spear thistle\end{center}&
        \begin{center}29 artichoke\end{center}\\
        
        $D^-_{KS}$&
        \begin{center}$-0.9517$\end{center}&
        \begin{center}$-0.9373$\end{center}&
        \begin{center}$-0.9293$\end{center}&
        \begin{center}$-0.9466$\end{center}&
        \begin{center}$-0.9438$\end{center}\\
        
    \end{tabular}
    \vspace{-10pt}
    \caption{Example of filters having an extremely negative $D_{KS}$ value on different classes of the \textit{flowers102} dataset and \texttt{VGG19} architecture. First row illustrates the visual pattern being captured by each feature, using the same method as in Figure \ref{fig:topKSimg_mit67}. The second and third row contain sample images from the \textit{flowers102} dataset for the different classes, and the name of those classes.}
    \label{fig:botKSimgflowers102}
\end{figure}

To illustrate the behavior of extremely negative $D_{KS}$ values, Figure \ref{fig:botKSimgcub200} shows a feature which has extremely negative $D_{KS}$ values (among the top 10 lowest) for four different classes of the \textit{cub200} dataset. This particular feature (the \texttt{n1946} of the \texttt{fc7} layer) is apparently specialized to recognize flying animals, as shown by the set of images from the \textit{ImageNet} validation set which produce a maximum feature activation (see first row of Figure \ref{fig:botKSimgcub200}).
A deeper analysis of the feature, based on the methods used by \shortciteR{yosinski2015understanding}, indicates that both the central colorful figure and the cluttered background are influential for the feature activation. Nevertheless, according to our $D_{KS}$ study, this feature produces top negative values for several classes of birds. The explanation behind this lies in the particularities of the classes for which the feature produces extremely negative $D_{KS}$ values: the four classes correspond to birds which live in a water or coastal environment, and which have dull colors (see second row of Figure \ref{fig:botKSimgcub200}). The feature, on the other hand, is apparently specialized on identifying colorful flying animals lying on branches. In this case, the extremely negative $D_{KS}$ values for this neuron would be analogous to identifying \textit{flying animals of dull colors} through the absence of visual features. Another example of this behavior for the flowers102 dataset is shown in Figure \ref{fig:botKSimgflowers102}. Again, two features which produce top 10 negative $D_{KS}$ values seem to be very representative of the whole dataset (classes of flowers), but not so for a few specific classes. One feature encodes the visual patterns corresponding to radial orange and red patterns (feature \texttt{n1449} of \texttt{fc7}), while the other focuses on wide petals (feature \texttt{n3529} of \texttt{fc7}). Clearly, the classes of flowers with highly negative $D_{KS}$ values do not have these properties. Hence, through the abnormal absence of both of these features (\eg the \textit{spear thristle} class shown in Figure \ref{fig:botKSimgflowers102}), we are roughly characterizing \textit{flowers without radial pistils and wide petals}.

These two examples illustrate how the lack of feature activations can convey relevant information. Notice how the behavior of negative $D_{KS}$ values depend on the context provided by the dataset, as extremely negative values on some classes will only happen for features which have a consistently high value on the rest of the dataset. Statistically, the extremely negative values of a feature can only happen for a small set of classes, since, if the set of classes grew, the inner/outer class disparity would decrease, making $D_{KS}$ closer to zero. This capability of extracting knowledge from the lack of data is novel and particularly relevant for feature representation transfer, where features are not originally designed for the target task. In this setting, both the presence and absence of visual patterns can provide relevant information for the characterization of images.

Let us also discuss the relevance of this behavior for fine-grained datasets, those containing classes belonging to a small, rather similar family of entities. Since extremely negative $D_{KS}$ values identify infrequently low feature activations, it is needed for that feature to be frequent on most of the dataset (\eg \textit{flying animals of bright colors that live on trees} is a frequent feature of birds). This may often happen in fine-grained datasets, where there are many common features in the data. However, in broad datasets which include a wider visual variety of classes (\eg \textit{ImageNet}, \textit{mit67}), there are much fewer features which are frequent on most classes and infrequent in a few. Hence, it will be much harder to obtain extremely negative $D_{KS}$ values.

\subsection{Statistic Distance Distributions}\label{sec:dist_dist}

\begin{figure}[b!]
    \centering
    \includegraphics[width=1\linewidth]{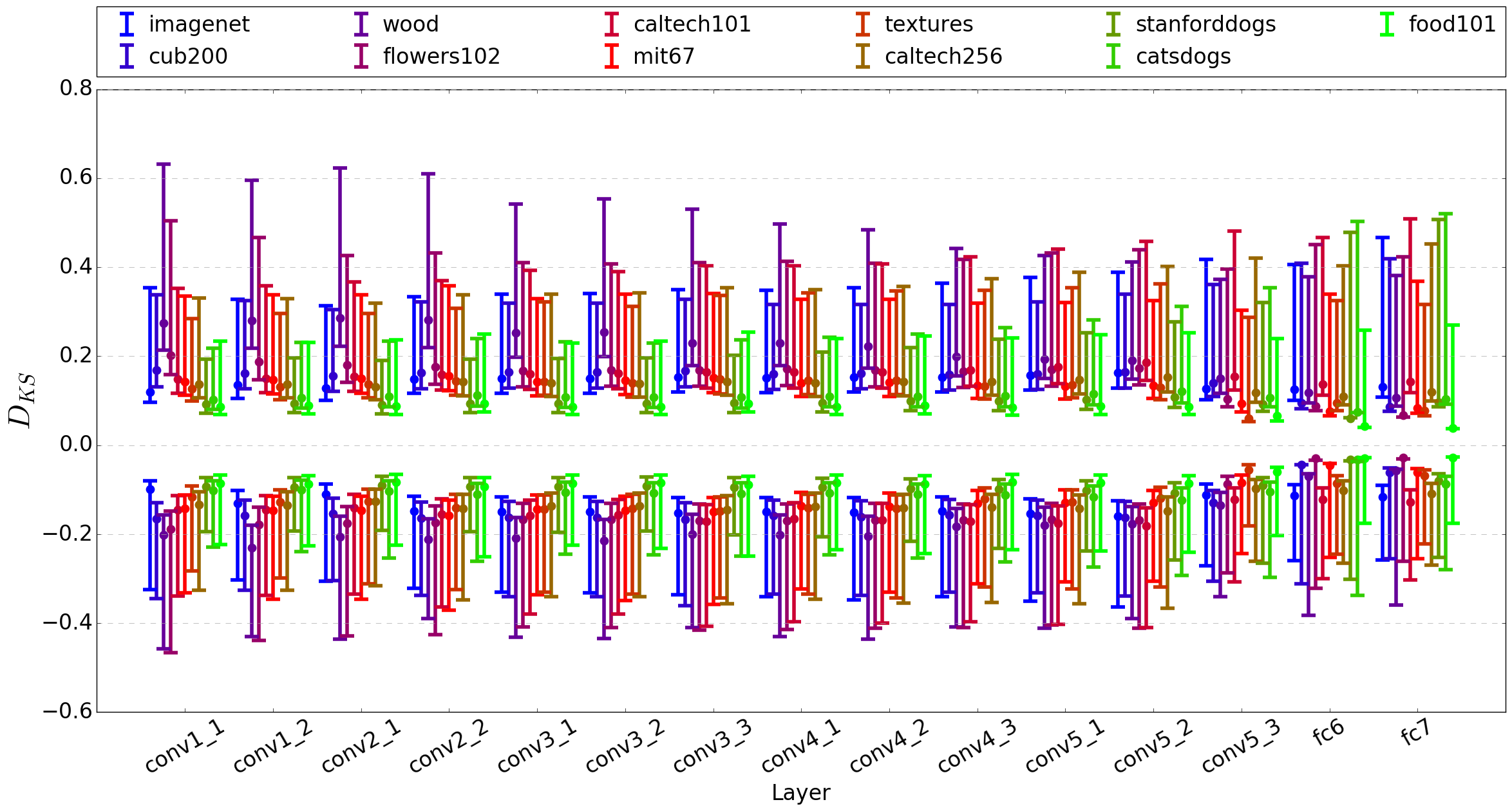}
    \vspace{-15pt}
    \caption{Inner/outer class $D_{KS}$ distribution per layer for 11 different datasets on the embedding. Since the distribution of $D^+_{KS}$ and $D^-_{KS}$ is similar to a log-normal distribution we represent it by the mode at the central point and two error bars enclosing 68\% of accumulated probability on each side (equivalent to $\mu \pm\sigma$ for normal distribution).}
    \label{fig:ks_dist_pl_3ds}
\end{figure}

After introducing the essential behavior of positive and negative $D_{KS}$, we now consider the overall distribution of $D_{KS}$ values per dataset. By plotting all $D_{KS}(f,c)$ values produced for a dataset, we obtain a clear bimodal distribution, separating positive ($D^+_{KS}$) and negative ($D^-_{KS}$) values. See Figure \ref{fig:ks_dist_1_53conv_vs_fc7_3ds} for an example. Each modality on its own resembles a log-norm distribution. To represent the distribution of $D_{KS}$ values for all layers and datasets in a single plot, Figure \ref{fig:ks_dist_pl_3ds} \textit{flattens} each distribution and displays the two corresponding modes and error bars. Before discussing the resultant distributions, let us define a few terms which we will use in the following sections.


A data representation which is good at modeling the target domain $\mathcal{D_T}$ can be considered to be highly descriptive, as it is capable of characterizing the associated \textit{data}. On the other hand, a data representation which is good at modeling the target task $\mathcal{T_T}$ can be considered to be highly discriminative, as it is capable of separating the associated \textit{labels}. This same categorization can be made for features, being highly descriptive the ones that help to build a rich representation of the domain, and highly discriminative the ones that help to solve the classification task. In the context of our study, the discriminativeness of a feature $f$ \wrt a class $c$ is shown by how close to either -1 or 1 $D_{KS}(f,c)$ is (as discriminative features are expected to produce very different values for $I_c$ and $I_{\neg c}$). Unfortunately, the descriptiveness of a feature cannot be illustrated in terms of $D_{KS}$ values, as descriptiveness originates from the domain and not from the task labels (which is what $D_{KS}$ measures). All further references to the discriminativeness of a feature will refer to this definition.


Let us now discuss the distributions of $D_{KS}$ values shown in Figure \ref{fig:ks_dist_pl_3ds}. Regardless of the layer depth, most features are discriminative for most tasks (either positively or negatively) as there are very few $D_{KS}$ values close to zero. The separation between $D^+_{KS}$ and $D^-_{KS}$ decreases on deeper layers (the fully-connected ones), particularly for those tasks which differ the most from the source task. Since deeper features are more specialized for the source task, more of these features may turn out to be irrelevant for the new task, producing similar activation values for $I_c$ and $I_{\neg c}$, which in turn results in $D_{KS}$ values closer to zero. Indeed, the distance between $D^+_{KS}$ and $D^-_{KS}$ does not decrease for those datasets which are essentially a subset of the source task, such as \textit{imagenet}, \textit{caltech101} and \textit{caltech256}. The behavior of $D_{KS}$ distributions on fully-connected layers is further discussed in Section\ref{sec:afc}.

To further investigate the variable behavior of $D_{KS}$ distributions based on layer depth, Figure \ref{fig:ks_dist_1_53conv_vs_fc7_3ds} shows the distributions separated in two plots: one for features from convolutional layers and another one for features from fully connected layers. According to the top plot of Figure \ref{fig:ks_dist_1_53conv_vs_fc7_3ds}, almost all features from convolutional layers are equally discriminative for all datasets, even for the tasks which are a direct subset of the source task (\eg \textit{imagenet}). Furthermore, the number and degree of positively discriminative features is almost symmetrical to the number and degree of negatively discriminative features. This indicates that convolutional features contain a similar amount of information to be exploited from both modalities. The behavior of $D_{KS}$ distributions on convolutional layers is further discussed in Section\ref{sec:acl}.




Most of these insights are coherent with the findings in the state-of-the-art, indicating that features from high-level layer are more specific and discriminant, particularly for target tasks which are close to the source task \shortcite{azizpour2016factors}. However, our results indicate that features from low-level layers are more general and discriminant than originally considered. This opens the door to use them for knowledge representations purposes and related problems such as unsupervised learning.

\subsection{Desirable Distributions of $D_{KS}$ Values}\label{sec:desirable}

Before getting into the detailed analysis, let us consider what characterizes a useful feature from the perspective of $D_{KS}$ distributions. This will help motivate some of the conclusions we draw from the consequent analysis.

\begin{figure}[t!]
    \centering
    \includegraphics[width=0.9\linewidth]{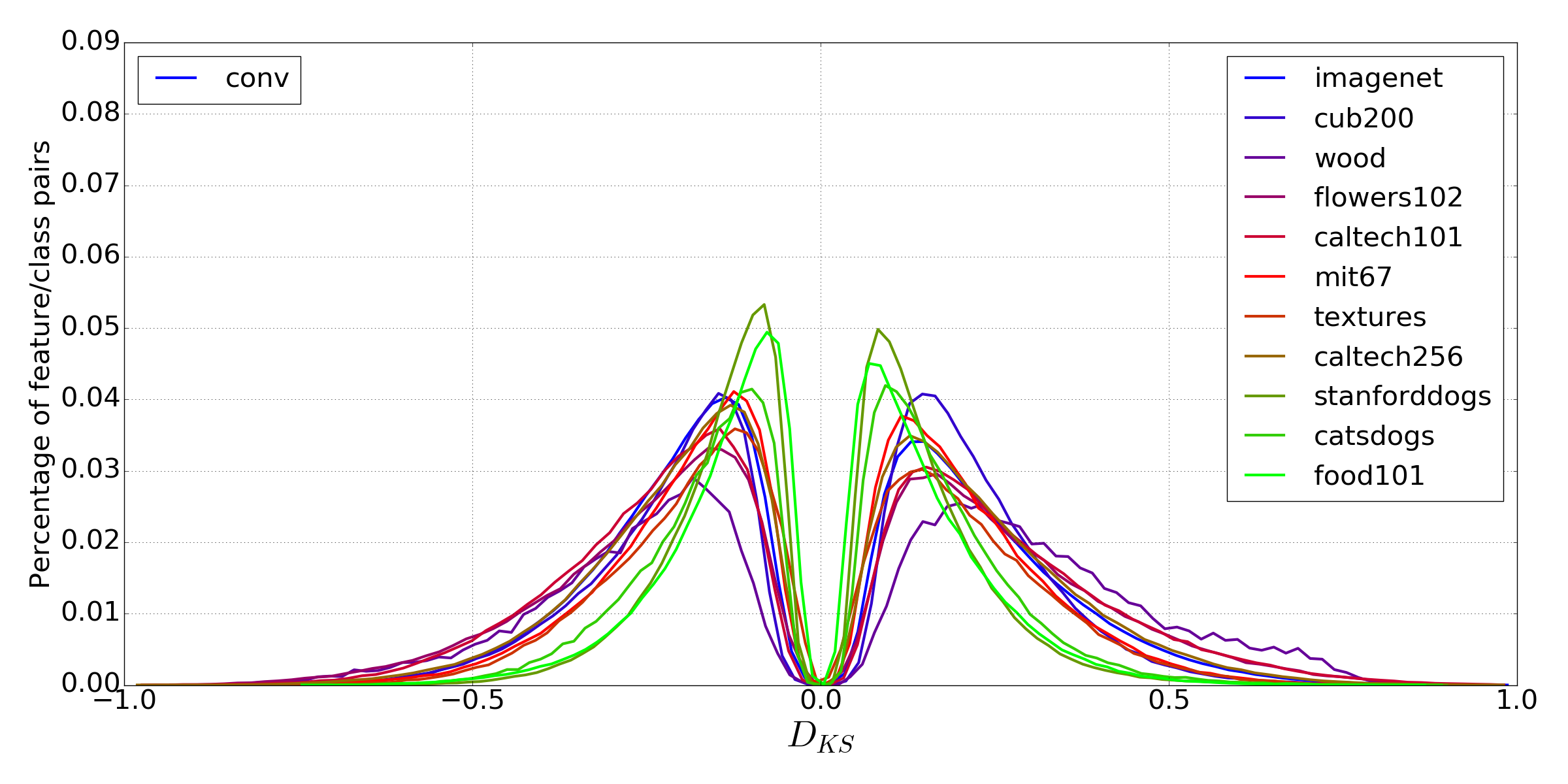}
    \includegraphics[width=0.9\linewidth]{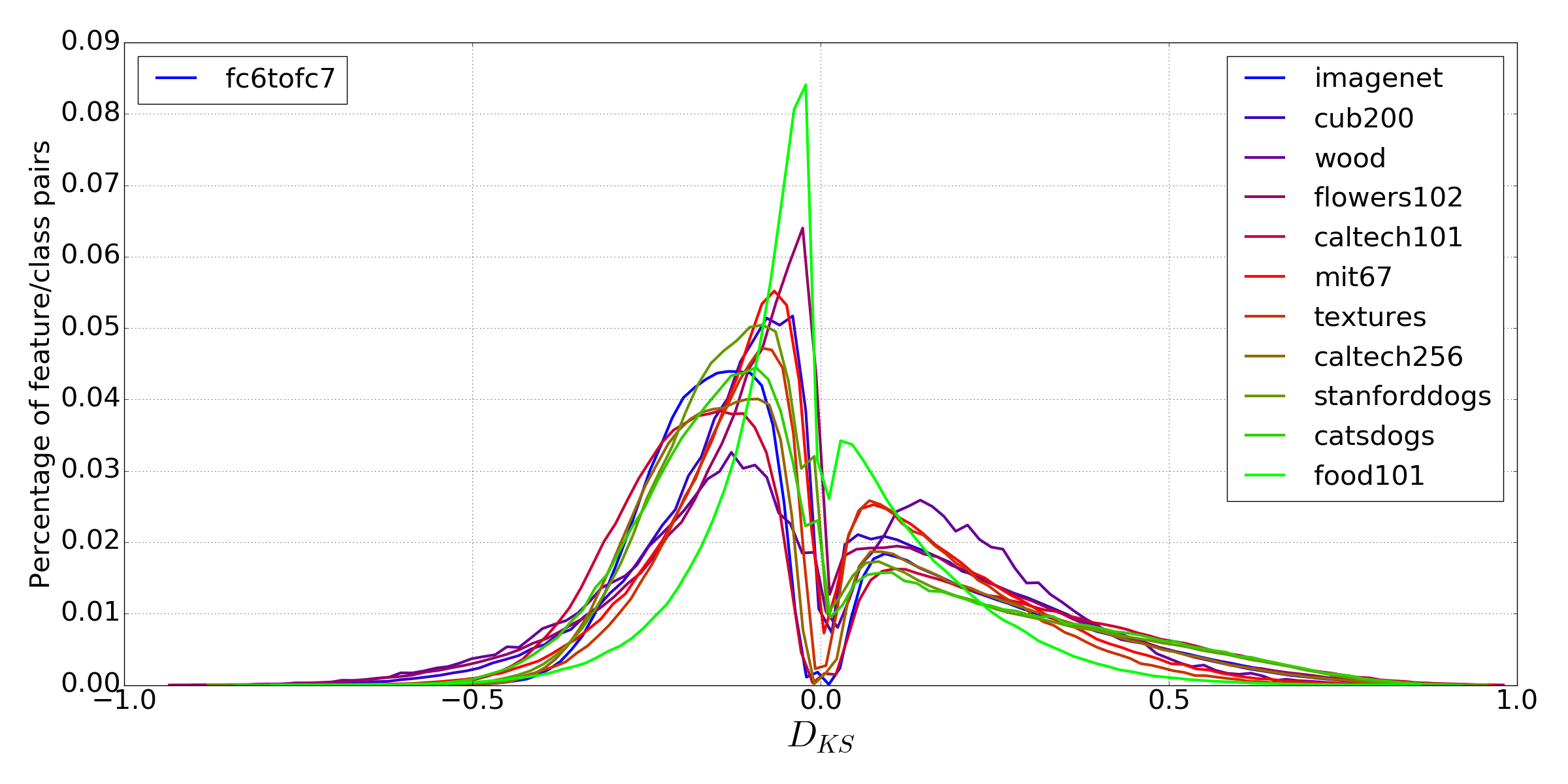}
    \vspace{-10pt}
    \caption{$D_{KS}$ distance distribution for convolutional (top) and fully connected (bottom) layers for 11 different datasets on the embedding. The x axis indicates the $D_{KS}$ values per feature and class, while the y axis indicates the occurrence as a percentage of the corresponding total features/class pairs $(f,c)$.}
    \label{fig:ks_dist_1_53conv_vs_fc7_3ds}
\end{figure}

As mentioned before, features with high absolute $D_{KS}$ value for a given task are discriminative of the classes of that task. Considering all the features of a layer together, as in Figure \ref{fig:ks_dist_1_53conv_vs_fc7_3ds}, the desirable distribution becomes one with density concentrated as much as possible on the extremes of the x axis. The two plots of Figure \ref{fig:ks_dist_1_53conv_vs_fc7_3ds} show that convolutional features are on average more separated from the irrelevancy of $D_{KS}=0$. However, in the bibliography there are plenty of experiments where a fully-connected layer is shown to outperform a convolutional layer for classification. The explanation for this phenomenon is that the correlation between $D_{KS}$ values and discriminativeness is not linear, as discriminativeness grows rapidly as it approaches $D_{KS}=1$ or $D_{KS}=-1$. As a result, having two features with $D_{KS}=0.3$ is not as good as having a single feature with $D_{KS}=0.5$. The higher discrimative power of fully-connected features reported in the bibliography for certain datasets is thus supported by the distribution of $D_{KS}$ values (bottom plot of Figure \ref{fig:ks_dist_1_53conv_vs_fc7_3ds}) which, for some datasets, has a slightly higher and longer tail on the plus side than the distribution for convolutional features (top plot of Figure \ref{fig:ks_dist_1_53conv_vs_fc7_3ds}). For other datasets (\eg \textit{wood}, \textit{flowers102}) convolutional features have more discrimative power than fully-connected features.


 \section{Analysis of Statistic Behaviors} \label{sec:sdb}

In this section we discuss some of the observations we make on the plots introduced in the previous section. In the two following sections we separately analyze the behavior of convolutional layers and fully-connected layers.

\subsection{Analysis of Convolutional Layers} \label{sec:acl}

In all Figures, datasets are ordered by average number of images per class. As shown in Figure \ref{fig:ks_dist_pl_3ds}, there is a clear correlation between that number and the $D_{KS}$ values obtained on the low convolutional layers (less images per class may cause more extreme $D_{KS}$ values). This correlation decreases with layer depth and is non-existent on the fully-connected layers. Middle and higher layers are more affected by other properties like dataset similarity, as we will see later.

\begin{figure}[t!]
    \centering
    \begin{tabular}{c *{4}{m{0.15\textwidth}}}
        &
        \multicolumn{4}{c}{35 alpine sea holly}\\
      \vspace{-5pt}
        &
        \multicolumn{4}{c}{\includegraphics[width=.20\linewidth]{flowers35}}\\
        \vspace{-15pt}
        Feature visualization&
        \begin{center}\includegraphics[width=1\linewidth]{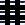}\end{center}&
        \begin{center}\includegraphics[width=1\linewidth]{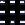}\end{center}&
        \begin{center}\includegraphics[width=1\linewidth]{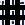}\end{center}&
        \begin{center}\includegraphics[width=1\linewidth]{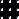}\end{center}\\ 
        \vspace{-10pt}
        neuron&
        \begin{center}\texttt{conv1\_2 n22}\end{center}&           
        \begin{center}\texttt{conv1\_2 n8}\end{center}&
        \begin{center}\texttt{conv1\_2 n16}\end{center}&
        \begin{center}\texttt{conv1\_1 n60}\end{center}\\
        
        $D^+_{KS}$&
        \begin{center}$0.8665$\end{center}&
        \begin{center}$0.8663$\end{center}&
        \begin{center}$0.8641$\end{center}&
        \begin{center}$0.8393$\end{center}\\

    \end{tabular}
    \vspace{-10pt}
    \caption{Example of low-level features (from layers \texttt{conv1\_1} and \texttt{conv1\_2}) with a high $D^+_{KS}$ value (among the top 10) for a given class of the flowers102 dataset on \texttt{VGG19} architecture. The first row contains a sample image of the corresponding class. The second row contain feature visualizations from \textit{imagenet}, obtained with the same process as in Figure \ref{fig:topKSimg_mit67}.}
    \label{fig:topKconv1flowers102}
\end{figure}

We start our analysis of the distributions for convolutional layers shown Figure \ref{fig:ks_dist_pl_3ds} by focusing on the unusually long bars for the \textit{wood} and, to a certain degree, also on the \textit{flowers102} datasets. This behaviour is more obvious in the first convolutional layers of the $D^+_{KS}$ modality, but becomes attenuated in later layers and is not symmetric for the $D^-_{KS}$ modality. Beyond having a relatively few images per class (a particular class of the \textit{wood} dataset has only 14 samples), the \textit{flowers102} dataset, and specially the \textit{wood} dataset, are composed by classes which differ only in small and texture-like characteristics. Low level convolutional layers are known to learn filters similar to Gabor filters and color blobs \shortcite{yosinski2014transferable}, which are appropriate to solve this sort of problems. These two factors explain why these features are so disproportionally discriminative for these datasets. As to why the \textit{textures} dataset does not display this behavior, when its a dataset specific of textural patterns, the answer lies in the composition of the dataset. In addition to having more images per class, images from the \textit{textures} dataset display textures at an image level, and by looking at a few pixels in the image (as low convolutional layers do) it is impossible to identify the texture (\eg there are large portions of images labeled as \textit{wrinkled} which do not show a single wrinkle). Coherently, the most discriminative features for this dataset are the ones found within middle and upper convolutional layers. As an example on the behavior of low level convolutional layers, Figure \ref{fig:topKconv1flowers102} shows some of the features from layers \texttt{conv1\_1} and conv1\_2 that produce very high $D^+_{KS}$ values for a specific class of the \textit{flowers102} dataset. These particular features correspond to horizontal gradients, vertical gradients and edge detectors, features which appear infrequently often in images of this class when compared to the rest of the flowers in the dataset.


Beyond the behavior of the \textit{wood} and \textit{flowers102} datasets for the first convolutional layers, the distribution of $D_{KS}$ values is rather stable in general. The top plot of Figure \ref{fig:ks_dist_1_53conv_vs_fc7_3ds} shows that low-level convolutional features behave similarly for all datasets (including \textit{imagenet}). Even though these features were optimized for the classification of \textit{ImageNet 2012} classes, it seems that they are roughly as discriminative for \textit{imagenet} as they are for the rest of datasets. This provides further evidence on why transfer learning for fine tuning produces such good results \shortcite{yosinski2014transferable}, but also indicates that features from these layers could be used almost ubiquitously for knowledge representation. The only dataset behaving clearly differently at the extremes values of low-level layers features (by having a higher tail) is the \textit{wood} dataset, for the reasons previously discussed: very detailed classes and small sample size. \textit{flowers102} and \textit{caltech101} also have tails with a height above average, as these datasets also include these properties (both in the case of \textit{flowers102}, and only limited samples sizes in the case of \textit{caltech101}).


\subsection{Analysis of Fully Connected Layers} \label{sec:afc}

Let us now consider the distributions of $D_{KS}$ values for the fully connected layers through the bottom plot of Figure \ref{fig:ks_dist_1_53conv_vs_fc7_3ds}. The dataset \textit{food101} has the most distinct distribution, with a large spike of $D_{KS}$ values close to 0 (close to 8\% of feature class pairs fall within the same bin) and very few $D_{KS}$ values close to both -1 and 1. This behaviour is likely to be directly related with the variability of the domain, as well as with the number of images per class (\textit{food101} has the most, with 250). For something as inconsistent as food, a large number of samples may lead to very different activations within a class (\eg a \textit{caesar salad} can include many different ingredients presented in many different ways). These variations lead to indistinguishable inner-class and outer-class behaviors, which in turn results in $D_{KS}$ values close to zero.

\begin{figure}[t!]
    \centering
    \begin{subfigure}[a]{0.87\textwidth}
        \centering
        \includegraphics[width=0.87\linewidth]{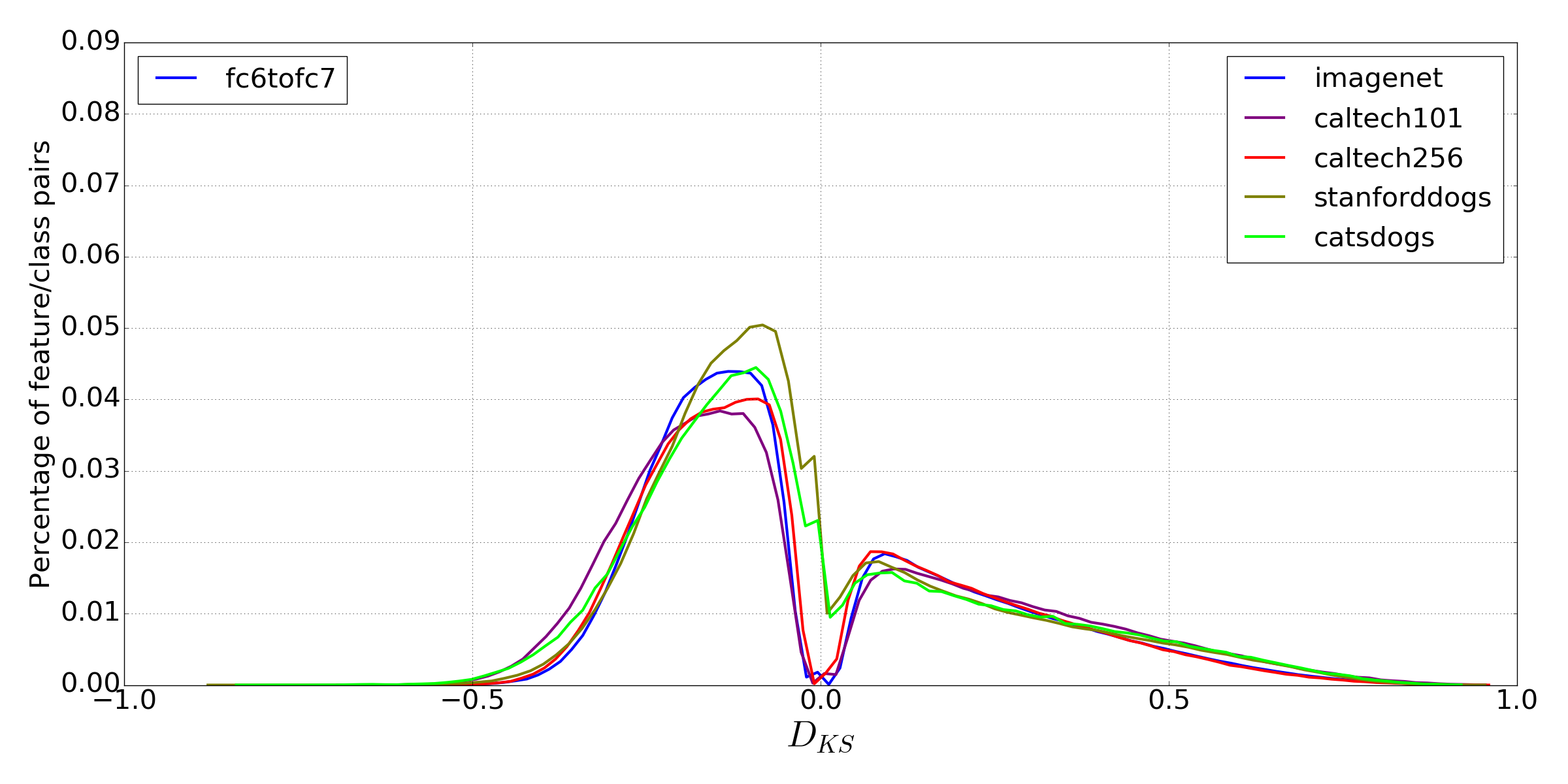}
        \caption{Datasets which are a subset of \textit{ImageNet2012}}
    \end{subfigure}
        \begin{subfigure}[b]{0.87\textwidth}
        \centering
        \includegraphics[width=0.87\linewidth]{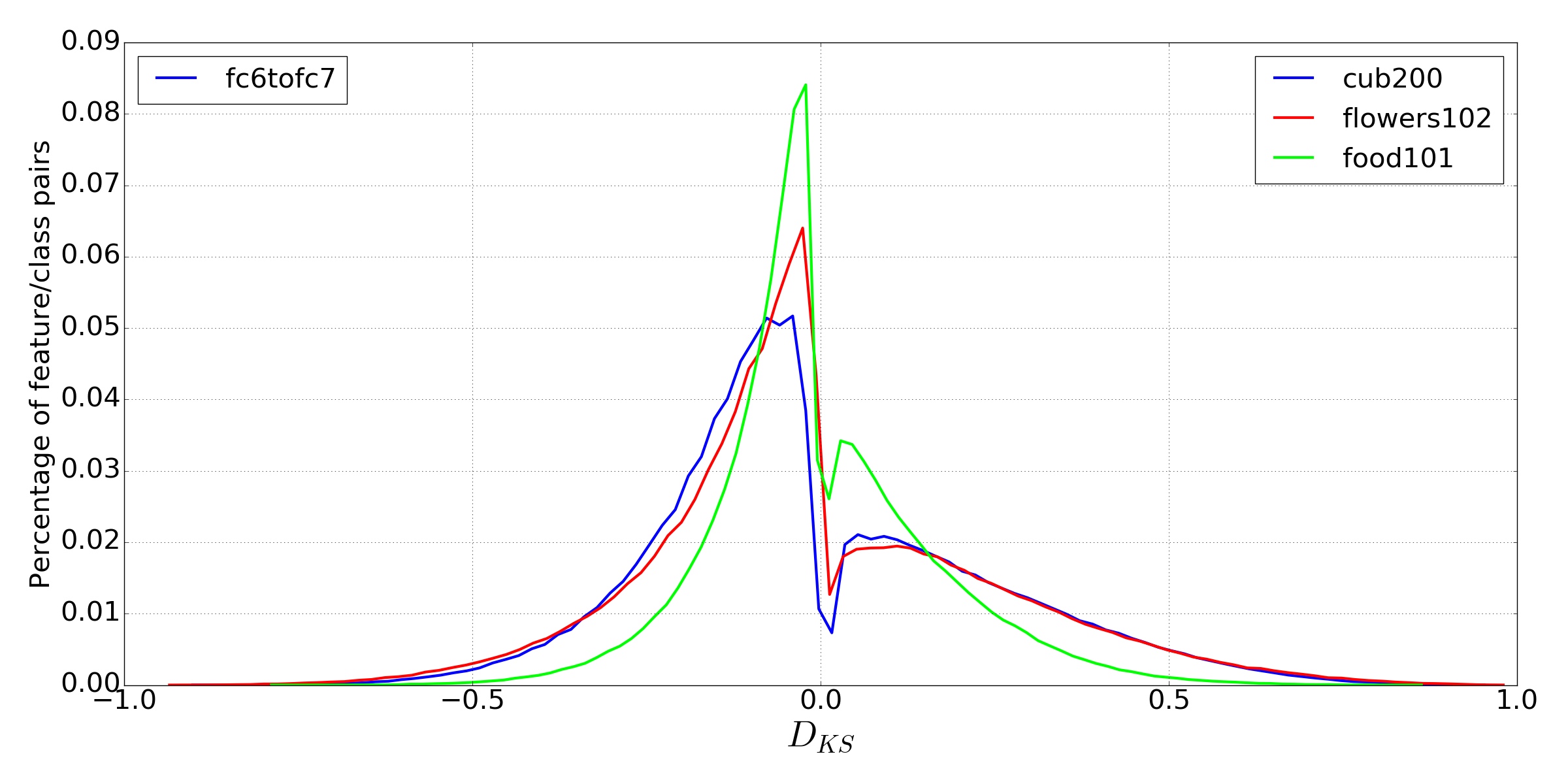}
        \caption{Datasets which intersect with \textit{ImageNet2012}}
    \end{subfigure}
        \begin{subfigure}[c]{0.87\textwidth}
        \centering
        \includegraphics[width=0.87\linewidth]{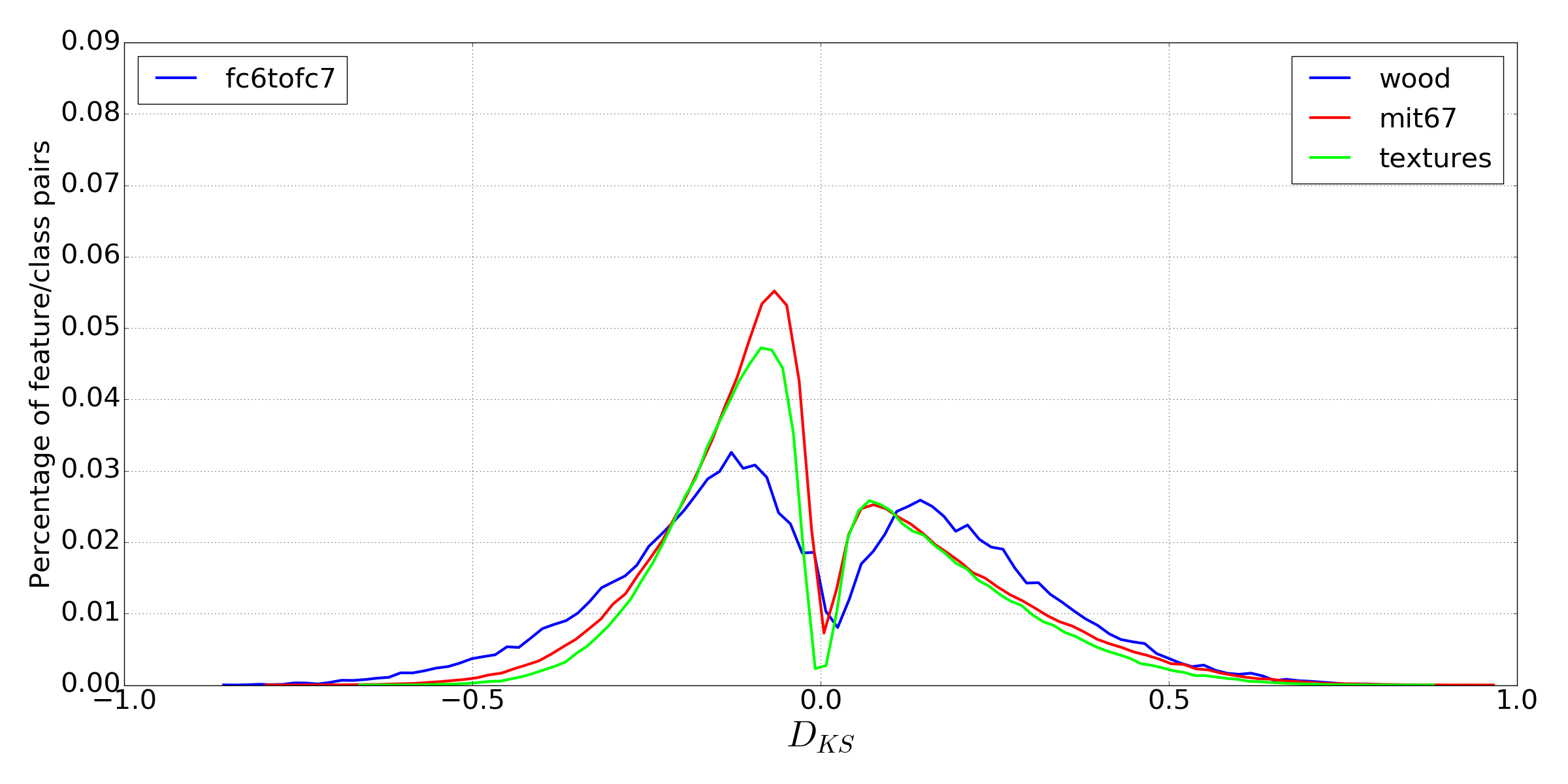}
        \caption{Datasets which are disjoint with \textit{ImageNet2012}}
    \end{subfigure}

    \vspace{-10pt}
    \caption{Distribution of $D_{KS}$ values for the features from the fully connected layer. Each plot show a subset of datasets, grouped by their similarity with \textit{ImageNet2012}. The x axis indicates $D_{KS}$ values, while the y axis indicates the occurrence as a percentage of the total.}
    \label{fig:ks_dist_fc_groups}
\end{figure}

The particular behavior of the \textit{food101} dataset does not extrapolate to the other datasets which also have a larger number of images per class, such as \textit{catsdogs} \textit{stanforddogs} and \textit{caltech256}. Since these datasets are very similar to the source task $\mathcal{T_S}$ (\textit{ImageNet2012}), these results indicate that similarity between tasks is the most relevant property for the behavior of fully-connected features. This is further supported by the distributions corresponding to the \textit{caltech101} and \textit{caltech256} datasets. While there are differences in their average number of instances per class $\overline{|I_c|}$ (91 - 120), their total size (9,146 - 30,607) and the number of classes (101 - 256) both $D_{KS}$ distributions are almost identical (see plot (a) of Figure \ref{fig:ks_dist_fc_groups}). To explore this consideration, next we categorize the 11 datasets in 3 groups, based on their degree of overlap with \textit{ImageNet2012}:

\begin{enumerate}[(a)]
    \item Datasets where the classes are a direct subset of the ones in \textit{ImageNet2012}. This group includes \textit{imagenet}, \textit{stanforddogs}, \textit{catsdogs}, \textit{caltech101} and \textit{caltech256}.
    \item Datasets where the classes partially intersect with the ones in \textit{ImageNet2012}. This group includes \textit{cub200}, \textit{flowers102} and \textit{food101}.
    \item Datasets where the classes are completely disjoint with the ones in \textit{ImageNet2012}. This group includes \textit{wood}, \textit{mit67} and \textit{textures}.
\end{enumerate}



Figure \ref{fig:ks_dist_fc_groups} shows the distribution of $D_{KS}$ values for the fully-connected features, plotted separately for each of these three groups. Group (a) is the only group where the distribution of $D_{KS}$ values gets very close to zero in the y axis for $D_{KS}=0$ for three of the five datasets in the group. This implies that, in these three datasets, there is not a single feature-class pair which has an identical inner and outer-class distribution. In other words, for these datasets all fully-connected features are at least somewhat discriminative for all classes. The three datasets showing this behavior are \textit{imagenet}, \textit{caltech101} and \textit{caltech256}. All wide spectrum datasets directly contained within the source task of \textit{ImageNet2012}. Significantly, this happens regardless of the number of classes (1,000, 101 and 256 respectively). On the other hand, the two datasets of group (a) for which this does not happen (\textit{stanforddogs} and \textit{catsdogs}) are limited to a certain domain (dogs, and cats and dogs respectively). Even though these datasets are subsets of the source task, there are still some fully-connected features which are not discriminant for any class. These irrelevant feature class pairs most likely correspond to those features used to characterize the type of elements which are found in \textit{ImageNet2012} but not in these restricted domains datasets (\eg those used to characterize non-living things). These indiscriminant features prevent the $D_{KS}$ distribution to reach zero on the $D_{KS}=0$ point. The impact of including a wide spectrum of classes \wrt not having indiscriminant features is further supported by the dataset with the fourth lowest percentage of feature/class pairs close to $D_{KS}=0$. That is the \textit{textures} dataset (see panel (c) of Figure \ref{fig:ks_dist_fc_groups}), which, although apparently has little in common with the source \textit{ImageNet2012} task, includes a wide variety of textures coming from plants, animals, man-made objects, \etc

In general, the bimodal distributions for both groups (a) and (b) are more imbalanced than for group (c), as the $D^-_{KS}$ part of the distribution accounts for a  significantly larger proportion of the total area. On the other hand, the distribution of values for the group (c) is closer to the distribution of values for the convolutional features (see top plot of Figure \ref{fig:ks_dist_1_53conv_vs_fc7_3ds}), where both modalities are symmetrical. The imbalanced behavior on groups (a) and (b) is explained by the same nature of fully-connected features, which were optimized during its original training to strongly activate for a small subset of classes and to be inhibited for the vast majority. This also results in a higher tail on the $D^+_{KS}$ side. On the other hand, the more balanced behavior of group (c) indicates that in this cases, instead of activating very strongly for a few set of classes, features activate moderately for a larger amount of classes. This is particularly interesting, as it indicates that fully-connected features could be treated as convolutional features when the target task is completely different than the source task.

\section{Level of Noise and Thresholding}\label{sec:apdpc}

In Section \ref{sec:sdb}, we discussed the distribution of $D_{KS}$ values at a dataset level, assuming that the $D_{KS}$ values were evenly distributed among the classes that compose a dataset. However, this may not be the case, as a subset of the classes composing a dataset may have a large set of relevant features, while another subset of classes is under-represented with no or very few features characterizing them. To answer this question, in Figure \ref{fig:acum_ks_dummy} we plot an accumulated distribution of $D^+_{KS}$ values per class. Each of the black lines represents a single class in the dataset. This graph is accumulative, showing how many features have a $D^+_{KS}$ value greater than the $x$ axis value for each class. Thus, at $D_{KS}=0.2$ (on the $x$ axis) we are plotting the number of features that meet $D_{KS}>0.2$ (on the $y$ axis) for each class.

\begin{figure}[t!]
    \centering
    \setlength{\tabcolsep}{0pt}
    \begin{tabular}{c c}
        \multicolumn{2}{c}{\textit{imagenet}} \vspace{-2pt}\\
        \multicolumn{2}{c}{\includegraphics[width=0.5\linewidth]{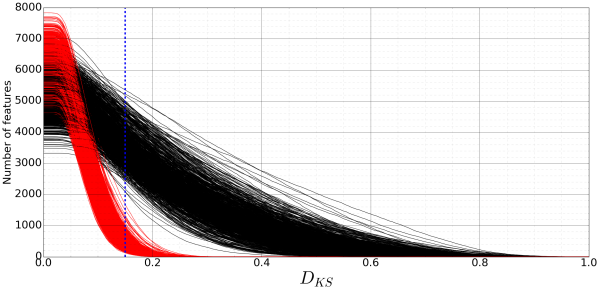}} \vspace{-5pt}\\

        \textit{cub200} & 
        \textit{wood} \vspace{-2pt}\\
        \includegraphics[width=0.5\linewidth]{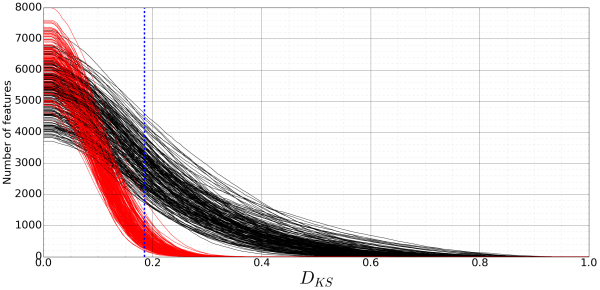} &
        \includegraphics[width=0.5\linewidth]{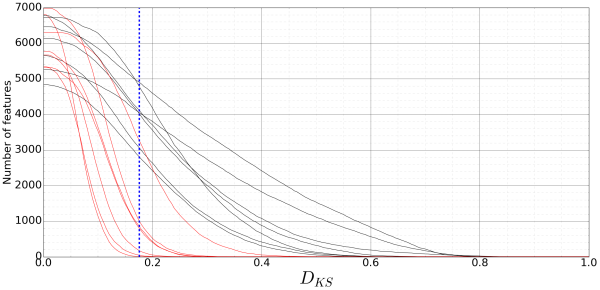}\\        
        
        \textit{flowers102} & 
        \textit{caltech101} \vspace{-2pt}\\
        \includegraphics[width=0.5\linewidth]{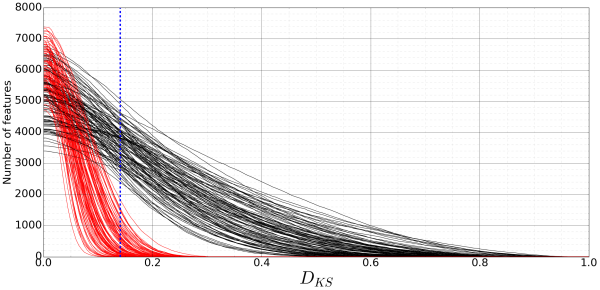} &
        \includegraphics[width=0.5\linewidth]{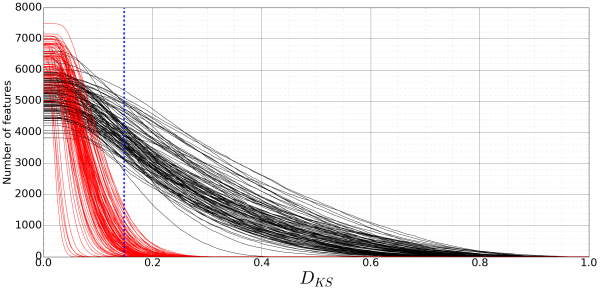}\\        
        
        \textit{mit67} & 
        \textit{textures} \vspace{-2pt}\\
        \includegraphics[width=0.5\linewidth]{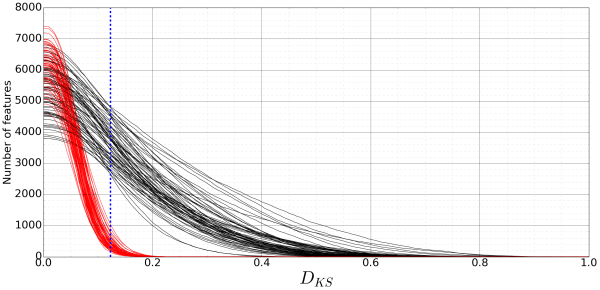} &
        \includegraphics[width=0.5\linewidth]{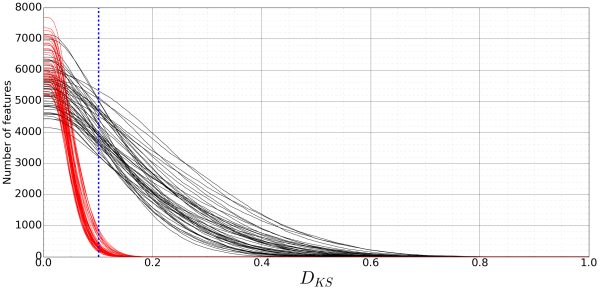}\\

    \end{tabular}
    \vspace{-10pt}
    \caption{$D_{KS}$ positive distance ($D^+_{KS}$) accumulated distribution for 11 different datasets (black) on the embedding. Each line corresponds to a different class of the dataset. Results for the same dataset with randomized labels are shown in red. Dashed line marks the $t^+$ threshold introduced in Section \ref{sec:pfa}.}
    \label{fig:acum_ks_dummy}
\end{figure}

\begin{figure}[t!]\ContinuedFloat
    \centering
    \setlength{\tabcolsep}{0pt}
    \begin{tabular}{c c}
        \textit{caltech256} & 
        \textit{stanforddogs} \vspace{-2pt}\\
        \includegraphics[width=0.5\linewidth]{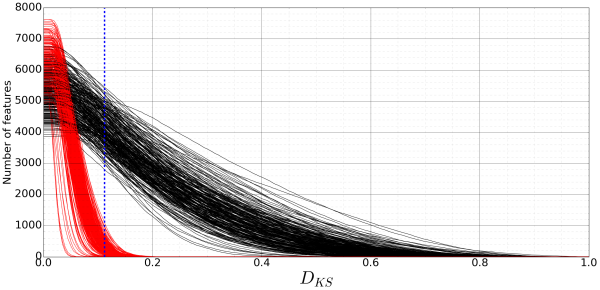} &
        \includegraphics[width=0.5\linewidth]{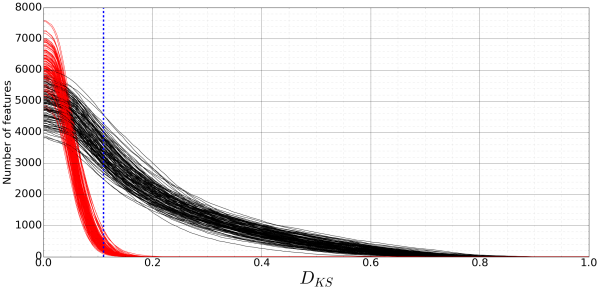}\\        
        
        \textit{catsdogs} & 
        \textit{food101} \vspace{-2pt}\\
        \includegraphics[width=0.5\linewidth]{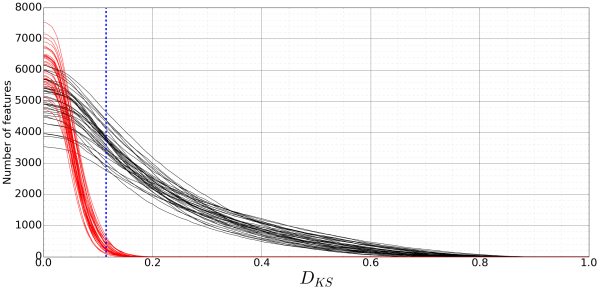} &
        \includegraphics[width=0.5\linewidth]{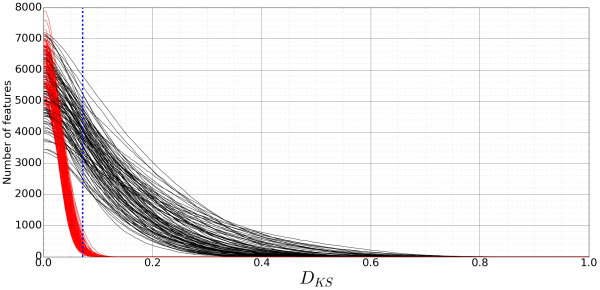}\\
    \end{tabular}
    \vspace{-10pt}
    \caption{(cont.) $D_{KS}$ positive distance ($D^+_{KS}$) accumulated distribution for 11 different datasets (black) on the embedding. Each line corresponds to a different class of the dataset. Results for the same dataset with randomized labels are shown in red. Dashed line marks the $t^+$ threshold introduced in Section \ref{sec:pfa}.}
    \label{fig:acum_ks_dummy2}
\end{figure}

Figure \ref{fig:acum_ks_dummy} shows a certain variance among classes of the same dataset for any given $D_{KS}$ threshold. This implies that some classes are more richly characterized by the embedding than others, as suspected. Although no class reaches $0$ on the $y$ axis until around $D_{KS}=0.4$, some reach $0$ at a remarkable $D_{KS} = 0.9$. To check if having $0$ features at $D_{KS}=0.4$ implies that the class is characterized by the embedding, in the same figure we show the behavior of the same dataset with randomized labels (in red). This is obtained assigning to each image a random label, keeping the total number of instances per class unmodified (\ie shuffling the real labels). Notice that this process keeps unchanged properties like the number of classes or the imbalance in the number of instances per class. By randomizing the labels we can observe the characterization that the embedding produces of purely noisy classes with the same characteristics of the target task.

As shown for all 11 datasets of Figure \ref{fig:acum_ks_dummy}, most randomized classes drop to 0 features between $D_{KS} = 0.1$ and $D_{KS} = 0.3$. The gap between the black and red lines allows us to assert that all classes are represented meaningfully (\ie beyond randomness) at a certain point. It also triggers the question of which portion of this curve could or should be pruned to maximize discriminativeness while minimizing noise. This is equivalent to ask which is the minimum $D_{KS}$ value we consider to be relevant when choosing the features to characterize a class.

\subsection{Threshold Measure} \label{sec:pfa}

One of the main goals of this paper is to study the viability of using convolutional features for feature representation transfer. However, and due to their unspecificity, many convolutional features may generate noise in the sense that they do not provide any information related to the target labels. The plots of Figure \ref{fig:acum_ks_dummy} showing the inner and outer class distributions for randomized classes provides a first insight on the actual magnitude of that noise.


We now consider the definition of thresholds $t^+$ and $t^-$ on $D_{KS}$, such that every $D^+_{KS}<t^+$ or $D^-_{KS}>t^-$ could be safely discarded in a feature representation transfer process. These thresholds should allow us to determine which features are likely to be relevant for each class, canceling out a significant amount of noise. Defining such a threshold implies a trade-off, as a $t^+$ and $t^-$ close to zero would result in representations with a larger descriptive power, while a $t^+$ and $t^-$ close to 1 or -1 respectively would result in representations with a minimum amount of noise.
 
As a reliable threshold (this is analogous for both $t^+$ and $t^-$), we propose one which maximizes the distance between a datasets and its corresponding version with randomized labels (as shown in each subfigure of Figure \ref{fig:acum_ks_dummy}). We define such distance using the average number of features having a $D_{KS}>x$ for all x in the range $[0,1]$. This is analogous to compute, for every point along the $x$ axis of one of the subfigures of Figure \ref{fig:acum_ks_dummy}, the average $y$ axis values for all black/red lines. The average of black lines will give us the behavior on the regular dataset, while the average of red lines will give us the behavior on its randomized version. By obtaining the difference between both values we obtain the \textit{average distance} ($d_{avg}$) between a dataset and its randomized version. Formally, the average distance for a value $D_{KS}=x$ is:
 
\begin{equation} \label{eq:avgdist}
    \begin{split}
    d_{avg}(x) = \frac{\sum_{ c\in C}|D_{KS}(f,c)>x|}{|C|} - \frac{\sum_{c'\in C_{rand}}|D_{KS}(f,c')>x|}{|C_{rand}|} \\ 
    \forall f \in embedding
    \end{split}
\end{equation}

where $C$ are the known classes (labels) associated with the data, and $C_{rand}$ are the randomly associated classes (random-labels). The vertical bars $|\cdot|$ denote set cardinality. 

\begin{table}[b!]
\centering
\caption{$t^+$ and $t^-$ thresholds as defined by the maximum average distance for each of the eleven datasets explored. $D^+_{KS}$ and $D^-_{KS}$ regions are computed separately. The third and fifth column shows the maximum $d_{avg}$ distance between the dataset and its randomized version.}
\label{tab:DKSthreshold}
\begin{tabular}{lcccc}
    \toprule
    Dataset &  $t^+$ & $d_{avg}(t^+)$  & $t^-$ & $d_{avg}(t^-)$  \\
    \midrule
    \textit{imagenet}       &   0.150   &   3,100    &   -0.121      & 4,368             \\
    \textit{cub200}         &   0.185   &   2,289    &   -0.155      & 2,417           \\
    \textit{wood}           &   0.176   &   3,087    &   -0.156      & 2,901           \\
    \textit{flowers102}     &   0.141   &   3,186    &   -0.120      & 3,572           \\
    \textit{caltech101}     &   0.148   &   3,480    &   -0.124      & 4,737           \\
    \textit{mit67}          &   0.123   &   3,269    &   -0.106      & 3,745           \\ 
    \textit{textures}       &   0.101   &   3,738    &   -0.089      & 4,553           \\
    \textit{caltech256}     &   0.112   &   3,813    &   -0.096      & 5,172           \\
    \textit{catsdogs}       &   0.115   &   3,353    &   -0.098      & 4,727           \\
    \textit{stanforddogs}   &   0.110   &   3,055    &   -0.092      & 4,535           \\
    \textit{food101}        &   0.072   &   3,555    &   -0.062      & 4,075           \\
    \bottomrule
\end{tabular}
\end{table}

Given the distance measure $d_{avg}(x)$, we define the thresholds $t^+$ and $t^-$ as the values of $D_{KS}=x$ that maximize $d_{avg}(x)$ in each respective sub-domain $D^+_{KS}$ and $D^-_{KS}$. Table \ref{tab:DKSthreshold} shows the thresholds found for the 11 datasets.

 

There is a clear correlation between $t^+$ and $t^-$ and the number of samples per class $\overline{|I_c|}$. Indeed, a logarithmic curve can be fitted to both $t^+$ and $t^-$ with respect to $\overline{|I_c|}$ obtaining coefficients of determination $R^2$ of 0.82 and 0.84 respectively. This indicates that the number of images per class is a very good indicator of the level of noise to be expected. This factor overshadows other relevant aspects, such as the level of similarity between tasks, which is only important when the tasks are exactly the same (\ie \textit{imagenet}).

\begin{figure}[b!]
    \centering
    \setlength{\tabcolsep}{0pt}
    \begin{tabular}{c c c c }

        \textit{caltech101} &
        \textit{caltech256} &
        \textit{stanforddogs} &         
        \textit{catsdogs}
        \vspace{-2pt}\\
        
        \includegraphics[width=0.24\linewidth]{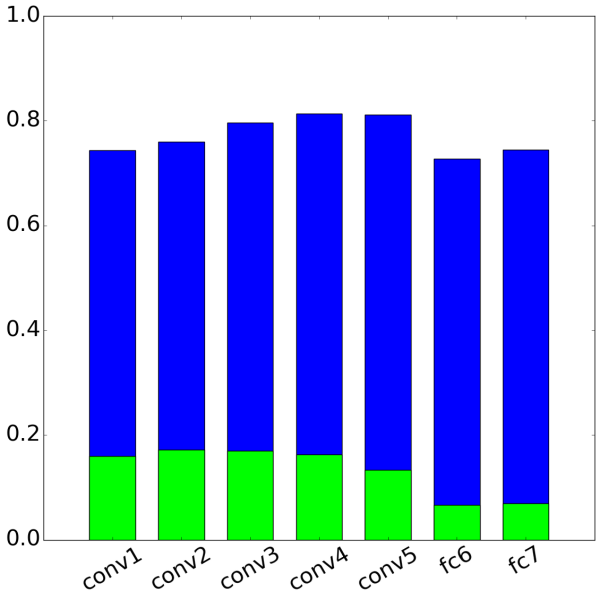} &
        \includegraphics[width=0.24\linewidth]{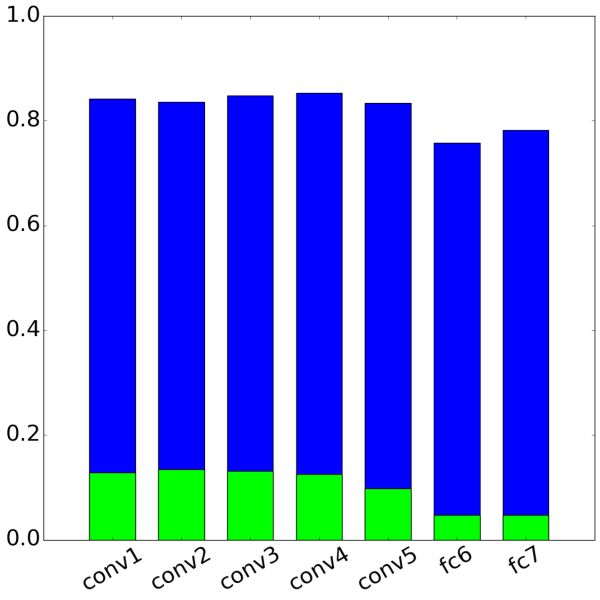} &
        \includegraphics[width=0.24\linewidth]{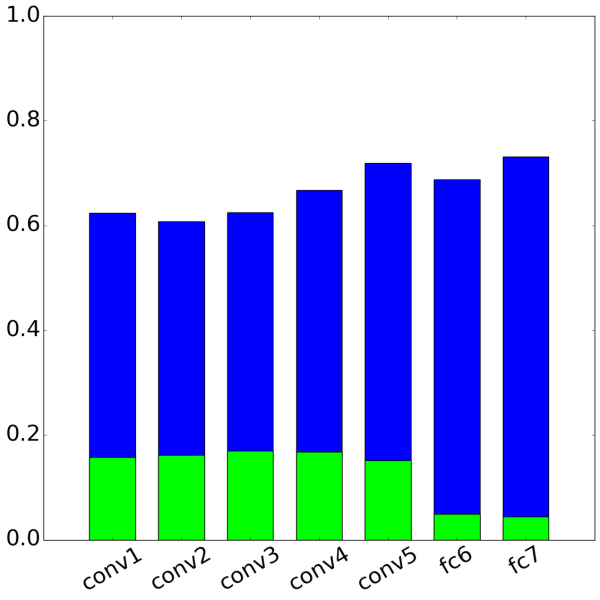} &
        \includegraphics[width=0.24\linewidth]{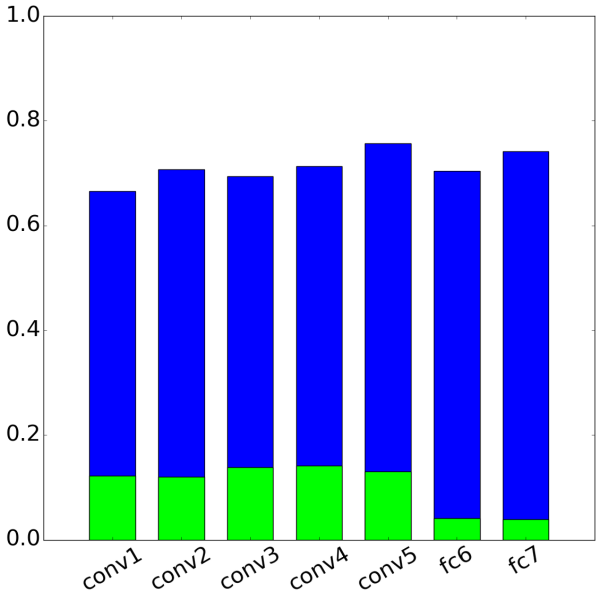} \\

        \textit{imagenet} &
        \textit{cub200} &
        \textit{wood} &
        \textit{flowers102} 
        \vspace{-2pt}\\
        
        \includegraphics[width=0.24\linewidth]{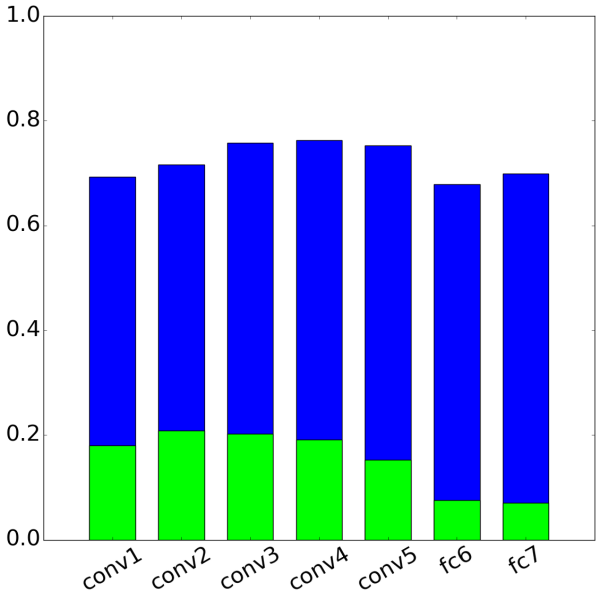} &
        \includegraphics[width=0.24\linewidth]{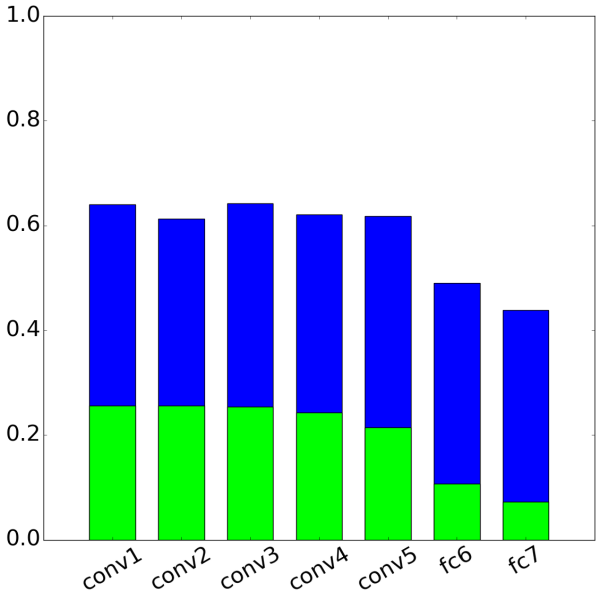} &
        \includegraphics[width=0.24\linewidth]{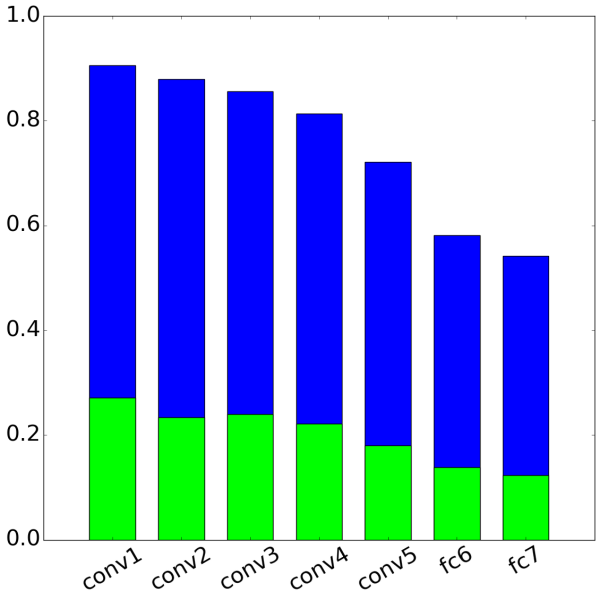} &
        \includegraphics[width=0.24\linewidth]{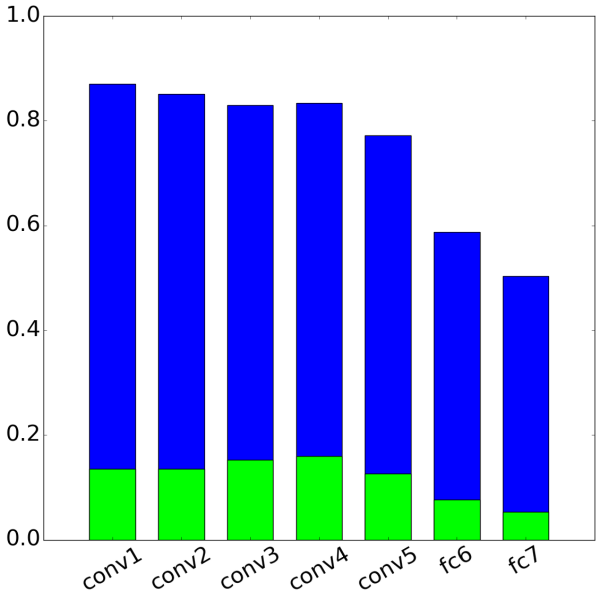}\\

        \textit{mit67} &
        \textit{textures} &
        \textit{food101} 
        \vspace{-2pt}\\
         
        \includegraphics[width=0.24\linewidth]{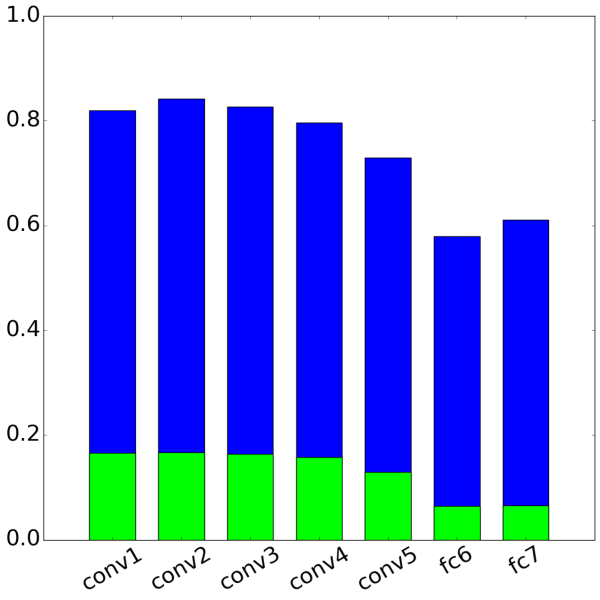} &
        \includegraphics[width=0.24\linewidth]{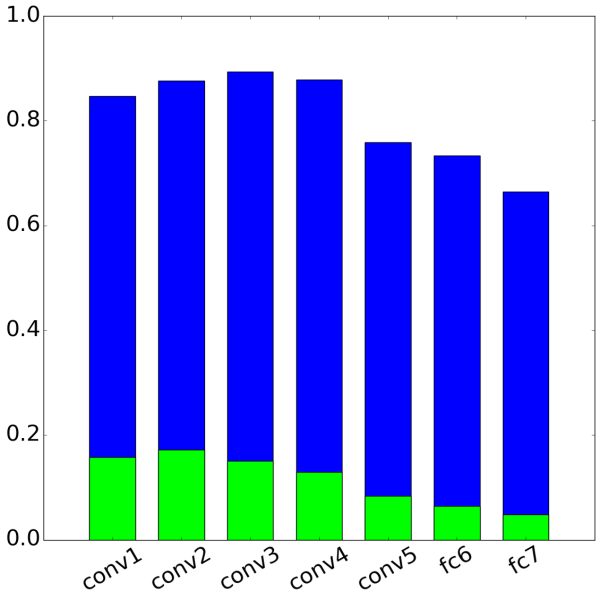} &
        \includegraphics[width=0.24\linewidth]{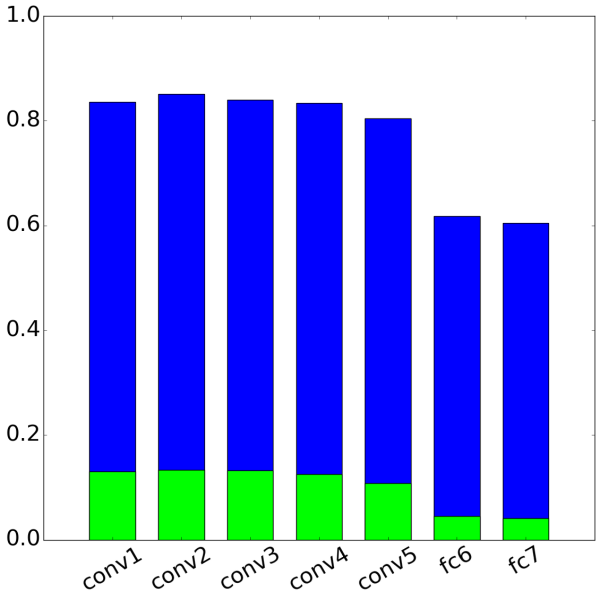}\\
        
    \end{tabular}
    \vspace{-10pt}
    \caption{For each dataset and layer, percentage of feature/class pairs remaining after pruning, for the original labels (blue) and for the randomized labels (green) (\ie features whose $D_{KS}$ for a certain class is higher than $t^+$ or lower than $t^-$).}
    \label{fig:pruned_dataset_by_layers}
\end{figure}

It is also interesting to see how many features in the embedding remain relevant after pruning the noisy ones through the application of the threshold. Of the 12,416 features in the embedding, approximately 3,300 features for $t^-$ and 4,000 for $t^+$ remain on average. The threshold corresponding to each dataset is plotted as a vertical dashed line in Figure \ref{fig:acum_ks_dummy}, showing how all classes (all black lines) would be minimally represented after applying it.



To study the degree of noise layer-wise, in Figure \ref{fig:pruned_dataset_by_layers} we plot the percentage of $D_{KS}(f,c)$ values that are kept by the $t^+$ and $t^-$ thresholds on various sets of layers. For datasets in group (a) (\ie  \textit{imagenet}, \textit{stanforddogs}, \textit{catsdogs}, \textit{caltech101} and \textit{caltech256}), the pruned feature-class pairs are evenly distributed among convolutional and fully-connected layers. This indicates that noise is found throughout the embedding for these datasets. For datasets in groups (b) and (c), the pruned feature-class pairs from fully-connected layers are significantly larger than the pruned pairs from convolutional layers. This is caused by the higher specificity of high level features, which are more frequently irrelevant for characterizing classes which differ from the source task. This results could be useful for, given a target task, determining which features and from which layers should be used when building an embedding.




\section{Consistency Between Source Tasks}\label{sec:places}

In this section we validate that our results are consistent beyond the source problem used for CNN training. For that purpose we use of the same VGG16 network architecture trained for the \textit{Places2} dataset \shortcite{zhou2016places}. \textit{Places2} is a task $\mathcal{T_S}$ unrelated to \textit{ImageNet 2012} containing a large set of samples (1.8 million). However, the domain $\mathcal{D_S}$ of \textit{Places2} dataset is not as wide as \textit{ImageNet 2012}, as it is focused in scene categories instead of objects.

\begin{figure}[b!]
    \centering
    \includegraphics[width=1\linewidth]{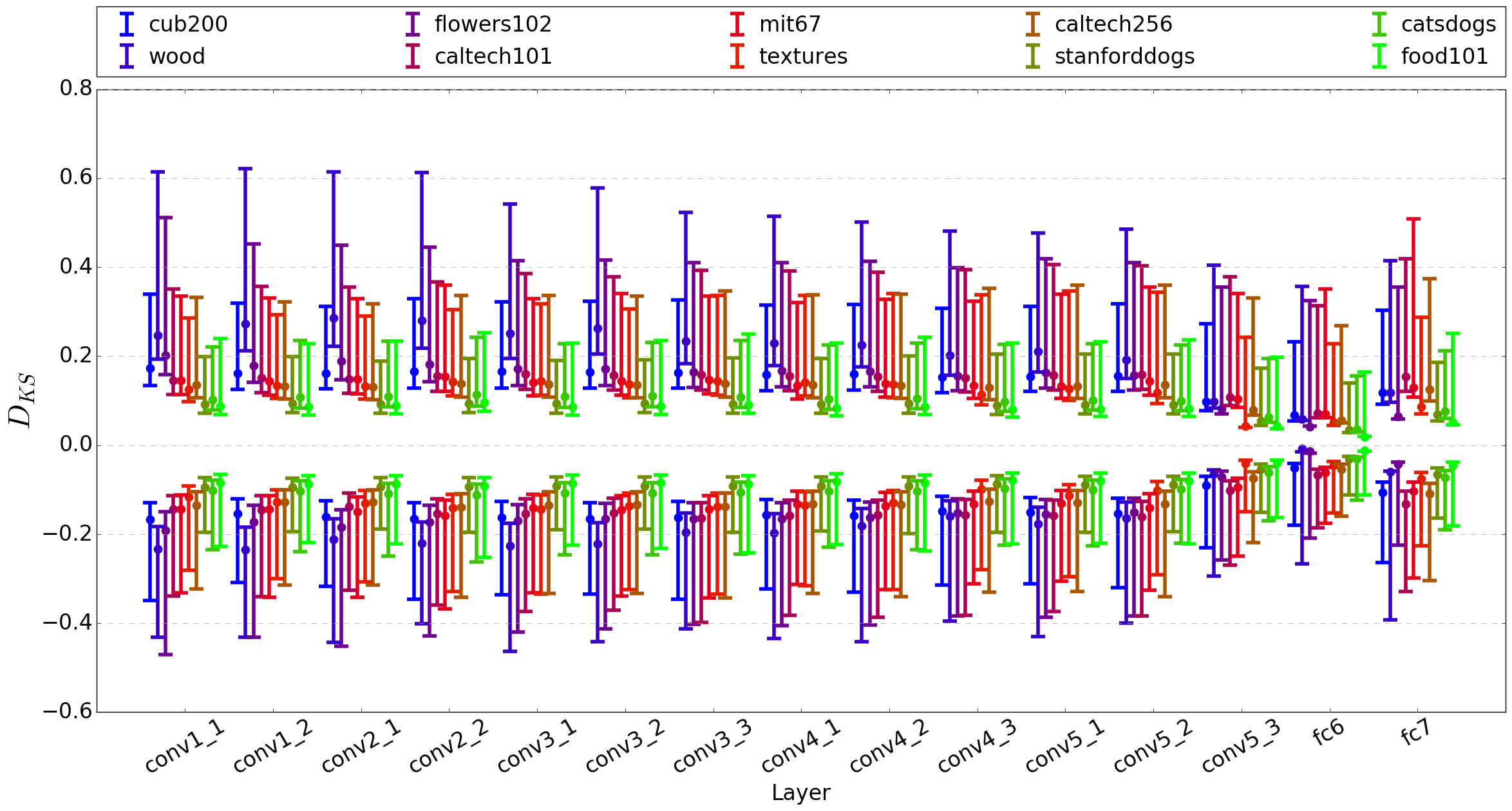}
    \vspace{-15pt}
    \caption{Inner/outer class $D_{KS}$ distribution per layer for 10 different datasets on the embedding created from \textit{places2} source task. Details are as in Figure \ref{fig:ks_dist_pl_3ds}}
    \label{fig:ks_dist_pl_places2}
\end{figure}

Analogous to Figure \ref{fig:ks_dist_pl_3ds}, Figure \ref{fig:ks_dist_pl_places2} shows the distribution of $D_{KS}$ values per layer using the embedding created by \textit{Places2} dataset. Overall, the distribution is quite similar to the one obtained with the \textit{ImageNet2012} embedding. Focusing on convolutional layers we confirm the observed correlation between the average number of instances per class $\overline{|I_c|}$ and $D_{KS}$ values. The particular behaviours of \textit{wood} and \textit{flowers102} is also present. Moreover, convolutional layers \texttt{conv1\_1} to \texttt{conv3\_3} look practically the same for all datasets. This similarity, reinforces the hypothesis of the generalist nature of convolutional features, regardless of the source and target tasks.

In the case of fully-connected features, the behavior of \textit{mit67} is analogous to those of group (a) for the \textit{ImageNet2012} embedding, as \textit{mit67} is now the closest task to the source task (\ie \textit{Places2}). Target tasks with no intersection with this source task, such as \textit{stanforddogs} and \textit{catsdogs}, now display a completely opposite activity.

There is however a remarkable difference between both embeddings in the second-to-last fully-connected layer, as the $D_{KS}$ divergences contract significantly for all datasets (including \textit{mit67}). Although we have no clear explanation for this phenomenon, we hypothesize that the different objects and characteristics needed to classify the holistic classes of \textit{Places2} (\ie scenes) cause the features from the \texttt{fc6} layer (where all this information is aggregated) to be extremely specific of the source task.

\section{Conclusions} \label{sec:c}

CNN feature representation transfer has been studied in the past through the performance of a classifier (most commonly, a SVM). Most contributions measure how each layer performs on its own at discriminating the classes of a task which is not the one the CNN was originally trained for. Through these contributions we know that, when considered together, the features composing a fully connected layer define the most discriminant of embedding spaces. In contrast with these contributions, the purpose of this paper was to analyze the behavior of all features from all layers individually, to measure their relevance for knowledge representation. We do so by exploring the inner/outer class activations of each feature, for all classes of several datasets. Some of the conclusions we draw from this study are coherent with the current state-of-the-art, and some are novel. Next we outline them all:

\begin{itemize}

\item Typically, features are characteristic for a given class by presence, but we have shown features can also be used to describe classes by their absence, thus providing a different type of information modality. This is particularly relevant for fine-grained datasets, where there may be many common features being characteristic for certain classes by their absence (\eg birds of dull colors that live in water). This novel contribution could be useful for knowledge representation and reasoning purposes (Section \ref{sec:sda}).
    
\item Features from the last convolutional layer and fully connected layers are highly specific, being either characteristic of a class or irrelevant for it. Features from the rest of convolutional layers convey more variate information, and can be characteristic of a class both by their presence or by their absence. This motivates the use of two distinct knowledge extraction approaches, depending on layer depth (Section \ref{sec:dist_dist}).

\item For certain tasks, convolutional features outperform fully-connected features at discriminating the corresponding labels. Overall, our results indicate that features from low-level layers are more general and discriminant than originally considered, and opens the door to use them for knowledge representations purposes and related problems such as unsupervised learning (Section \ref{sec:desirable}).

\item Low and middle level features have a very similar behavior for the dataset they were trained for (\textit{imagenet}) as for the rest of target datasets. This indicates that CNN features from these layers could be used for knowledge representation on a wide variety of datasets without fine-tuning. This is something previously proposed in the bibliography (Section \ref{sec:acl}).

\item As previously claimed in the bibliography, the relevance of fully-connected features is strongly related with the similarity of the task with the original problem the network was trained for. However, we find that wide spectrum tasks (\ie those containing all sorts of classes) is also a key factor. Only wide spectrum domains which are similar to the source task guarantee that there will be no indiscriminant fully-connected features (Section \ref{sec:afc}).

\item The behavior of fully-connected features for target tasks which have no intersection with the source task is similar to the behavior of convolutional features. In this context, both sets of features could be treated analogously for knowledge representation purposes (Section \ref{sec:afc}).

\item Discriminant features were found on all layers of the embedding for all classes of the eleven datasets evaluated. This means that, in a knowledge representation setting, no class would become indescribable, showcasing the richness of the representation language built by the CNN at every layer (Section \ref{sec:apdpc}).

\item Through the behavior of randomized datasets we obtain an estimation of the inner/outer class distances that can be accounted for by noise. We find that a conservative threshold with little variance can be defined across datasets, and after applying that threshold more than half of the features of the embedding remain relevant (Section \ref{sec:pfa}).

\item Context is key, both in the feature extraction and knowledge representation processes.
The significance of some feature activations (or its lack of) depends on the dataset being used as reference. The representation of data using neural network embeddings should consider context to be able to exploit all possible modalities of information.

\end{itemize}

Beyond these conclusions, this work provides a methodology for identifying relevant features (either by presence and absence) throughout a deep CNN. By applying this approach presented here, one can define a full-network embedding (an embedding using all layers of a network) which outperforms traditional single-layer embeddings in classification tasks \shortcite{garcia2017out}, and which improves the performance of multimodal pipelines for image caption and image retrieval tasks \shortcite{vilalta2017full}.


\acks{This work is partially supported by the Joint Study Agreement no. W156463 under the IBM/BSC Deep Learning Center agreement, by the Spanish Government through Programa Severo Ochoa (SEV-2015-0493), by the Spanish Ministry of Science and Technology through TIN2015-65316-P project and by the Generalitat de Catalunya (contracts 2014-SGR-1051), and by the Core Research for Evolutional Science and Technology (CREST) program of Japan Science and Technology Agency (JST).
}

\vskip 0.2in
\bibliographystyle{theapa}
\bibliography{biblio}

\end{document}